\pdfoutput=1
\documentclass[final,5p,times,twocolumn, number]{elsarticle}

\usepackage{graphicx}
\usepackage{graphics}
\usepackage{times}
\usepackage{amsmath}
\usepackage{amssymb}
\usepackage{float}
\usepackage{url}
\usepackage{subfigure}
\usepackage{caption}
\usepackage{multirow}
\usepackage{gensymb}
\usepackage{epstopdf}
\usepackage{wrapfig}
\usepackage{setspace} 
\usepackage[noend]{algpseudocode}
\usepackage{color}
\usepackage{booktabs}
\usepackage{multirow}

\journal{Arxiv}

\begin{document}

\begin{frontmatter}


\title{Title}
\author{Tixiao Shan$^{1}$, Jinkun Wang$^{2}$, Fanfei Chen$^{2}$, Paul Szenher$^{2}$, Brendan Englot$^{2}$}
\ead{shant@mit.edu, jwang92@stevens.edu, fchen7@stevens.edu, pszenher@stevens.edu, benglot@stevens.edu}
\address{$^{1}$Massachusetts Institute of Technology, 77 Massachusetts Ave, Cambridge, MA 02139, USA}
\address{$^{2}$Stevens Institute of Technology, 1 Castle Point Terrace, Hoboken, NJ 07030, USA}

\title{Simulation-based Lidar Super-resolution \\ for Ground Vehicles}


\author{}

\address{}

\begin{abstract}

We propose a methodology for lidar super-resolution with ground vehicles driving on roadways, which relies completely on a driving simulator to enhance, via deep learning, the apparent resolution of a physical lidar. To increase the resolution of the point cloud captured by a sparse 3D lidar, we convert this problem from 3D Euclidean space into an image super-resolution problem in 2D image space, which is solved using a deep convolutional neural network. By projecting a point cloud onto a range image, we are able to efficiently enhance the resolution of such an image using a deep neural network. Typically, the training of a deep neural network requires vast real-world data. Our approach does not require any real-world data, as we train the network purely using computer-generated data. Thus our method is applicable to the enhancement of any type of 3D lidar theoretically. 
By novelly applying Monte-Carlo dropout in the network and removing the predictions with high uncertainty, our method produces high accuracy point clouds comparable with the observations of a real high resolution lidar. We present experimental results applying our method to several simulated and real-world datasets. We argue for the method's potential benefits in real-world robotics applications such as occupancy mapping and terrain modeling.

\end{abstract}



\end{frontmatter}

\section{Introduction}
\label{sec::intro}

Light detection and ranging (lidar) is an essential sensing capability for many robot navigation tasks, including localization, mapping, object detection and tracking. Lidar uses light in the form of pulsed laser to measure relative range to surrounding objects. Unlike most cameras, which only function with sufficient ambient light, lidar will function even at night, offering long-range visibility and a wide horizontal aperture. 2D lidar is usually cost-efficient and has been widely used in many indoor applications such as mapping, localization, and obstacle avoidance. Recently, with the rapid development of self-driving vehicles, demand for 3D lidar has grown significantly. Though a revolving 2D lidar can mimic a 3D lidar by continuously changing the scanning position, such systems are often inefficient. A typical 3D lidar has multiple channels that revolve at different heights, producing a 3D point cloud with ring-like structure. The number of channels in the sensor determines the vertical density of its point clouds. A denser point cloud from a lidar with more channels can capture the fine details of the environment; applications such as terrain modeling and object detection can benefit greatly from a higher resolution lidar. However, increasing the number of channels can be very costly. For example, the most popular 16-channel lidar, the Velodyne VLP-16, costs around \$4,000. The 32-channel HDL-32E and Ultra Puck, and the 64-channel HDL-64E cost around \$30,000, \$40,000 and \$85,000 respectively.

In this paper, we propose what is to our knowledge the first dedicated deep learning framework for \textit{lidar super-resolution}, which predicts the observations of a high-resolution (hi-res) lidar over a scene observed only by a low-resolution (low-res) lidar. 
We convert the resulting super-resolution (super-res) point cloud problem in 3D Euclidean space into a super-res problem in 2D image space, and solve this problem using a deep convolutional neural network. Unlike many existing super-res image methods that use high-res real-world data for training a neural network, we train our system using only computer-generated data from a simulation environment. This affords us the flexibility to train the system for operation in scenarios where real hi-res data is unavailable, and allows us to consider robot perception problems beyond those pertaining specifically to driving with passenger vehicles.  
We investigate the benefits of deep learning in a setting where much of the environment is characterized by sharp discontinuities that are not well-captured by simpler interpolation techniques.
Furthermore, we use Monte-Carlo dropout \cite{dropout, yarin-gal-dropout} to approximate the outputs of a Bayesian Neural Network (BNN) \cite{BNN}, which naturally provides uncertainty estimation to support our inference task. 

\begin{figure*}[ht]
	\centering
	\includegraphics[width=.98\textwidth]{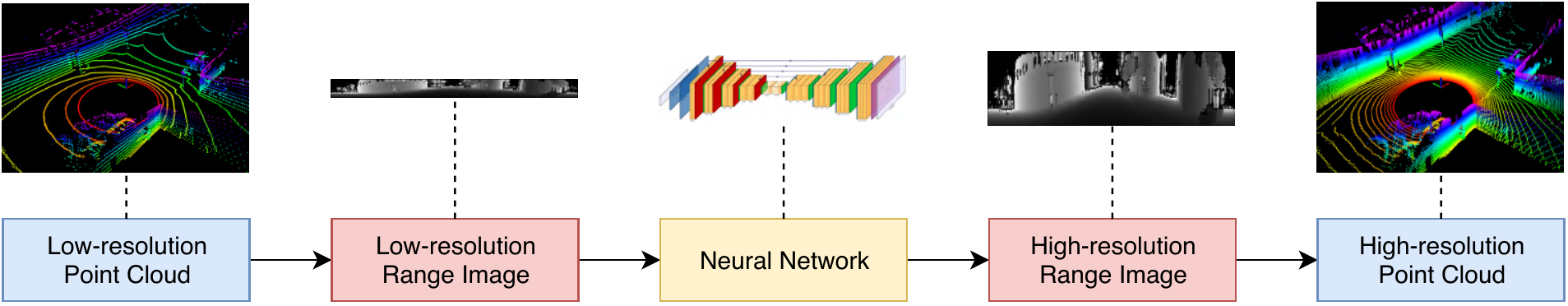}
	\caption{Workflow for lidar super-resolution. Given a sparse point cloud from a 3D lidar, we first project it and obtain a low-res range image. This range image is then provided as input to a neural network, which is trained purely on simulated data, for upscaling. A high-res point cloud is received by transforming the inferred high-res range image pixels into 3D coordinates.}
	\label{fig::process}
	\vspace{-3mm}
\end{figure*}

To the best of our knowledge, this is the first paper to present an approach for lidar super-resolution enabled by deep learning. It produces accurate high-res point clouds that predict the observations of a high-res lidar using low-res data. The contributions of this paper are as follows:
\begin{itemize} 
	\item A novel architecture for deep learning-enabled lidar super-resolution; 
	\item A procedure for training the architecture in simulation, which is thoroughly evaluated using datasets recorded in both simulated and real-world environments;
	\item A study of the framework's suitability for enhancing relevant robot mapping tasks, with comparisons against both deep learning and simpler interpolation techniques.
\end{itemize}


\section{Related Work}
\label{sec::related-work}

Our work is most related to the \textit{image super-resolution} problem, which aims to enhance the resolution of a low-res image. Many techniques have been proposed over the past few decades and have achieved remarkable results \cite{sr-review}. Traditional approaches such as linear or bicubic interpolation \cite{cubic-interpolation}, or Lanczos resampling \cite{Lcanzos}, can be very fast but oftentimes yield overly smooth results. Recently, with developments in the machine learning field, deep learning has shown superiority in solving many prediction tasks, including the image super-res problem. Methods based on deep learning aim to establish a complex mapping between low-res and high-res images. Such a mapping is usually learned from massive training data where high-res images are available. For example, a super-resolution convolutional neural network, SR-CNN, trains a three-layer deep CNN end-to-end to upscale an image \cite{sr-cnn-1}. Over time, deeper neural networks with more complex architectures have been proposed to further improve the accuracy \cite{SelfEXSR, DRCN, ESPCN, SR-GAN}. Among them, SR-GAN \cite{SR-GAN} achieves state-of-the-art performance by utilizing a generative adversarial network \cite{GAN}. The generator of SR-GAN, which is called SR-ResNet, is composed of two main parts, 16 residual blocks \cite{resnet} and an image upscaling block. A low-res image is first processed by the 16 residual blocks that are connected via skip-connections and then upscaled to the desired high resolution. The discriminator network of SR-GAN is a deep convolutional network that performs classification. It discriminates real high-res images from generated high-res images. It outperforms many other image super-res methods, including nearest neighbor, bicubic, SR-CNN and those of \cite{SelfEXSR, DRCN, ESPCN}, by a large margin.

Another problem that is related to our work is \textit{depth completion}. The goal of this task is to reconstruct a dense depth map with limited information. Such information usually includes a sparse initial depth image from a lidar or from an RGB-D camera \cite{rgb-d, ma-iros2016}. Typically, an RGB image input is also provided to support depth completion, since estimation solely from a single sparse depth image is oftentimes ambiguous and unreliable.
For instance, a fast depth completion algorithm that runs on a CPU is proposed in \cite{fast-cpu}. A series of basic image processing operations, such as dilation and Gaussian blur, are implemented for acquiring a dense depth map from a sparse lidar scan. Though this method is fast and doesn't require training on vast data, its performance is inferior when compared with many other approaches. 
A self-supervised depth completion framework is proposed in \cite{ma-arxiv-2018-july}. In this work, a deep regression network is developed to predict dense depth from sparse depth. The proposed network resembles an encoder-decoder architecture and uses sparse depth images generated by a lidar, with RGB images as optional inputs.
Another problem that is closely related to depth completion is \textit{depth prediction}, which commonly utilizes images from a monocular or stereo camera \cite{google-depth, depth-consistency, fully-conv, nvidia-depth}. Due to our focus here on a lidar-only super-resolution method, an in-depth discussion of this problem lies beyond the scope of this paper.

Instead of solving the super-resolution problem in image space, PU-Net \cite{pu-net} operates directly on point clouds for upsampling, and adopts a hierarchical feature learning mechanism from \cite{pointnet++}. However, this approach performs super-resolution on point cloud models of individual small objects, which differs from our approach that attempts to increase sensor resolution. The upsampled  high-res point clouds of our method retain the ``ring" structure characterizing the output of a real lidar. This preserves our approach's compatibility with other lidar perception algorithms that operate directly on 3D lidar scans as input.


\begin{figure*}[ht]
	\centering
	\subfigure[]{\includegraphics[width=.16\textwidth]{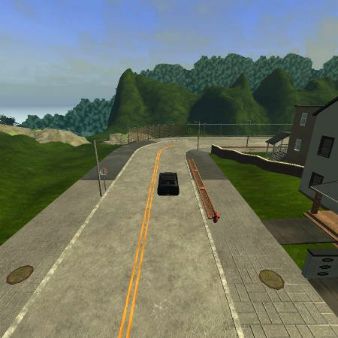}}
	\subfigure[]{\includegraphics[width=.16\textwidth]{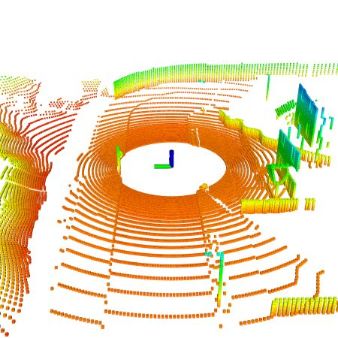}}
	\subfigure[]{\includegraphics[width=.16\textwidth]{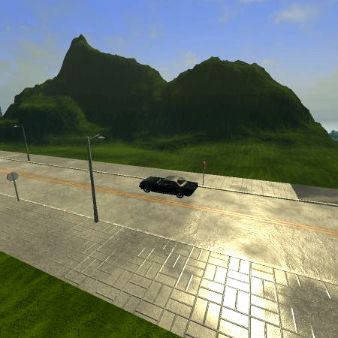}}
	\subfigure[]{\includegraphics[width=.16\textwidth]{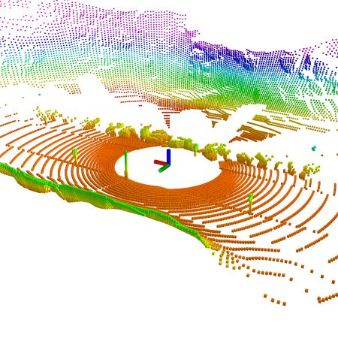}}
	\subfigure[]{\includegraphics[width=.16\textwidth]{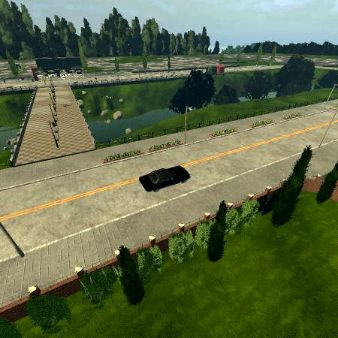}}
	\subfigure[]{\includegraphics[width=.16\textwidth]{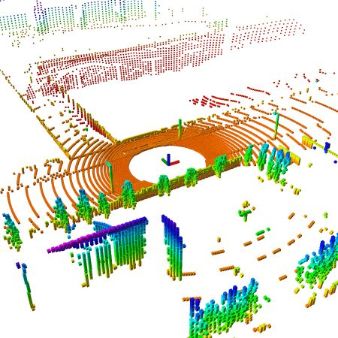}}
	\subfigure[]{\includegraphics[width=.16\textwidth]{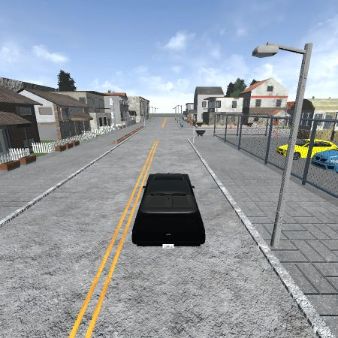}}
	\subfigure[]{\includegraphics[width=.16\textwidth]{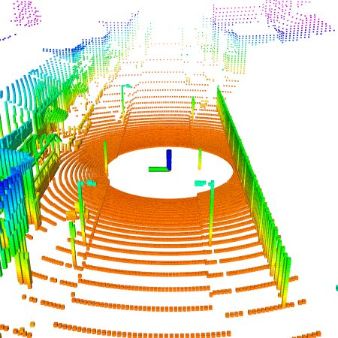}}
	\subfigure[]{\includegraphics[width=.16\textwidth]{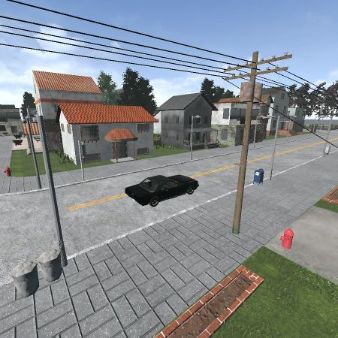}}
	\subfigure[]{\includegraphics[width=.16\textwidth]{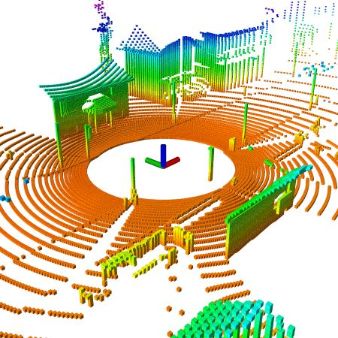}}
	\subfigure[]{\includegraphics[width=.16\textwidth]{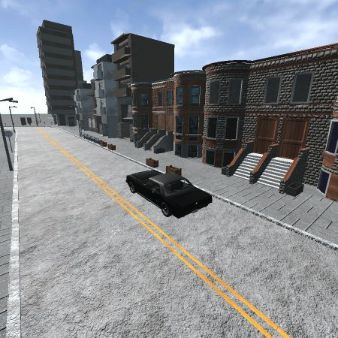}}
	\subfigure[]{\includegraphics[width=.16\textwidth]{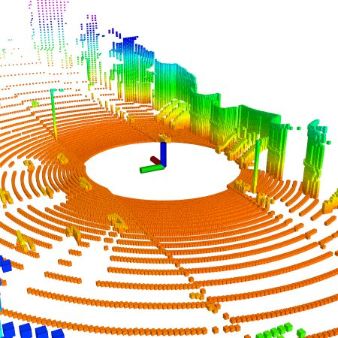}}
	\caption[CARLA Town 01 and 02 demonstration.]{Illustration of high-res point clouds (64-channel) captured in CARLA Town 01 (a-f) and Town 02 (g-l). CARLA Town 01 features a suburban environment with roads, trees, houses, and a variety of variable-height terrain. Town 02 features an urban environment.}
	\label{fig::carla-example}
	\vspace{-3mm}
\end{figure*}

\section{Technical Approach}

This section describes the proposed lidar super-resolution methodology in detail. Since the horizontal resolution of a modern 3D lidar is typically high enough, we only enhance vertical resolution throughout this paper. However, the proposed approach, without loss of generality, is also applicable for enhancing the horizontal resolution of a lidar with only a few modifications to the neural network. The workflow of the proposed approach is shown in Fig. \ref{fig::process}. Given a sparse point cloud from a 3D lidar, we first project it and obtain a low-res range image. This range image is then provided as input to a neural network, which is trained purely on simulated data, for upscaling. A dense point cloud is received by transforming the inferred high-res range image pixels into 3D coordinates.

\subsection{Data gathering}
\label{sec::data-gathering}

Similar to the method proposed in \cite{drive-matrix}, we leverage a rich virtual world as a tool for generating high-res point clouds with simulated lidar. There are many open source software packages, e.g. CARLA, Gazebo, Unity, that are capable of simulating various kinds of lidar on ground vehicles. Specifically, we opt to use the CARLA simulator \cite{carla} in this paper due to its ease of use and thorough documentation.

The first step involves identifying the lidar we wish to simulate. Let us assume we have a VLP-16 and we wish to quadruple (4$\times$ upscaling (16 to 64)) its resolution. The VLP-16 has a vertical field of view (FOV) of 30$^{\circ}$ and a horizontal FOV of 360$^{\circ}$. The 16-channel sensor provides a vertical angular resolution of 2$^{\circ}$, which is very sparse for mapping. We want to simulate a 64-channel ``VLP-64'' in CARLA, which also has a vertical and horizontal FOV of 30$^{\circ}$ and 360$^{\circ}$ respectively. With the simulated lidar identified, we can either manually or autonomously drive a vehicle in the virtual environment and gather high-res point clouds captured by this simulated lidar. An example of the lidar data produced in CARLA is shown in Figure \ref{fig::carla-example}.

We note that the simulated high-res lidar should have the same vertical and horizontal FOV as the low-res lidar. For example, we cannot train a neural network that predicts the perception of HDL-64E using the data from VLP-16 because their vertical FOVs are different.

\begin{figure*}[ht]
	\centering
	\includegraphics[width=.90\textwidth]{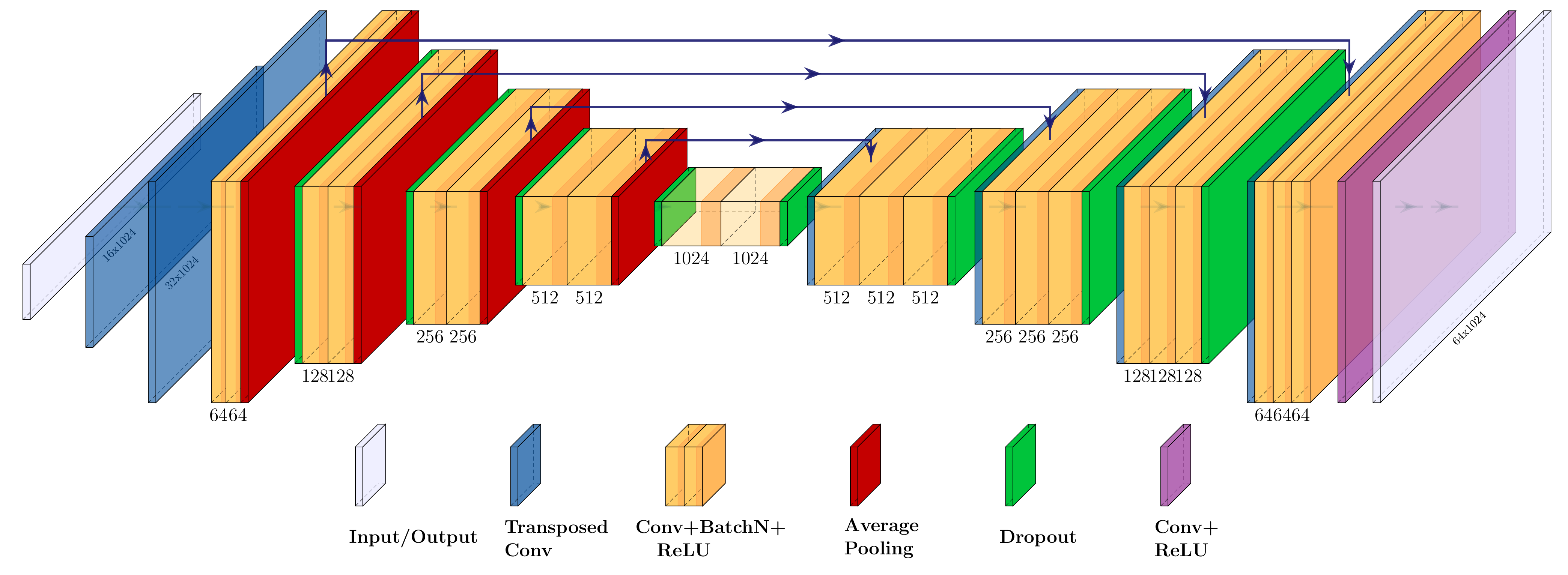}
	\caption{Our proposed neural network architecture for range image super-resolution. The network follows an encoder-decoder architecture. Skip-connections are denoted by solid lines with arrows.}
	\label{fig::unet}
	\vspace{-3mm}
\end{figure*}

\subsection{Data preparation and augmentation}
\label{sec::data-augmentation}

We then project the simulated high-res point cloud onto a range image, which can be processed by the neural network. A scan from the simulated ``VLP-64" 3D lidar will yield a high-res range image with a resolution of 64-by-1024. This high-res range image will serve as the ground truth comprising our training data. Then, we evenly extract 16 rows from this high-res range image and form a low-res range image, which has a resolution of 16-by-1024. This low-res range image is equivalent to the point cloud data captured by a VLP-16 after projection, and comprises the input to the neural network during training. We note that the resolution of the original range image from a VLP-16 sensor varies from 16-by-900 to 16-by-3600 depending on the sensor rotation rate. For the purpose of convenience and demonstration, we choose the horizontal resolution of all range images to be 1024 to accommodate different sensors throughout the paper. We also ``cut'' every range scan at the location facing the rear of the vehicle, for the purpose of converting it to a flattened 2D image. This is typically the region of least importance for automated driving and robot perceptual tasks, and is in many cases obstructed by the body of the vehicle. 

We then augment the data by performing top-down flipping, horizontal flipping and shifting, and range scaling to account for different environment structures and sensor mounting heights (such as driving on different sides of the road, and underneath structures).
To increase prediction robustness, we also vary sensor mounting attitudes during data gathering before augmentation.
Finally, the low-res and high-res range images are then normalized to $0-1$ and sent to train the neural network. For example, the maximum detection range of the VLP-16 is 100 meters. Thus we divide the ranges of the range image by 100 to obtain the normalized range image. For objects that are outside the sensor's range, their corresponding ranges in the image are set to be 0 as they yield no valid readings.


\subsection{Neural Network Architecture}
\label{sec::network}

The lidar super-res problem can now be formulated as an image super-res problem. Adapted from the encoder-decoder architecture of \cite{u-net}, we configure a neural network for range image super-resolution, shown in Figure \ref{fig::unet}. The input, low-res range image is first processed by two transposed convolutions for increasing the image resolution to the desired level. Then the encoder consists of a sequence of convolutional blocks and average pooling layers for downsampling the feature spatial resolutions while increasing filter banks. On the other hand, the decoder has a reversed structure with transposed convolutions for upsampling the feature spatial resolutions. All convolutions in the convolutional blocks are followed by batch normalization \cite{batch-norm} and ReLU \cite{relu}. The output layer produces the final high-res range image using a single convolution filter without batch normalization.

\begin{figure}[h]
	\centering
	\includegraphics[width=.9\columnwidth]{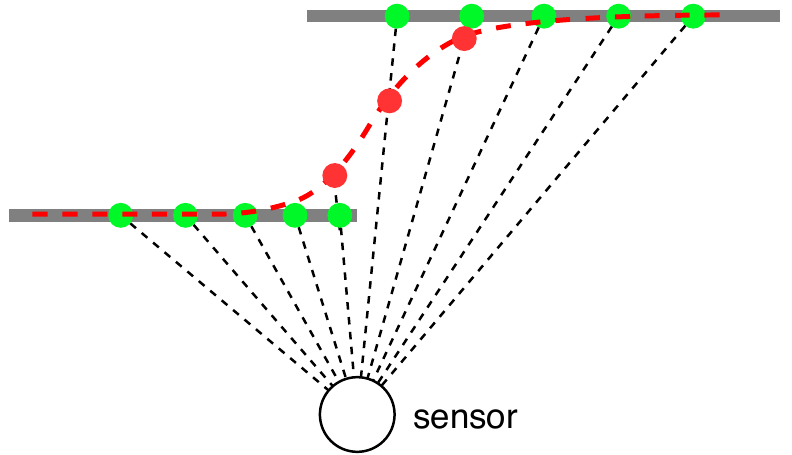}
	\caption{Smoothing effects after applying convolutions.}
	\label{fig::sensor-noise}
	\vspace{-6mm}
\end{figure}

\subsection{Noise Removal}
\label{sec::noise-removal}

We note that we have placed numerous dropout layers before and after the convolutional blocks in Figure \ref{fig::unet}. This is because performing convolutional operations on a range image will unavoidably cause smoothing effects on sharp and discontinuous object boundaries \cite{pseudo-lidar}. An illustrative example of this smoothing effect is shown in Figure \ref{fig::sensor-noise}. Ten range measurements from a lidar channel are shown in a top-down view. The gray lines represent two walls, and the green dots indicate the true range measurements. After convolution, the range measurements are smoothed (shown by the red curve) in places where environmental discontinuities occur. Incorporating smoothed range predictions, such as the three red dots shown, into a robot's point cloud will deteriorate the accuracy of the resulting map.

To address this problem, we novelly apply Monte-Carlo dropout (MC-dropout) for estimating the uncertainty of our range predictions \cite{yarin-gal-dropout}. MC-dropout regularization approximates a BNN by performing multiple feed-forward passes with active dropout at inference time to produce a distribution over outputs \cite{yarin-gal-dropout}. Given observations $\mathcal{D} = \{(\mathbf{x_i}, y_i)_{i=1:N}\}$, we seek to infer a probability distribution of a target value parameterized on the latent space $\Theta$: 
\begin{align}
	p(y^*|\mathbf x^*, \mathcal D) \propto \int p(y^*|\boldsymbol{\theta}^*)p(\boldsymbol{\theta}^*|\mathbf x^*, \mathcal D)d\boldsymbol{\theta}^*, 
	\label{eq::posterior}
\end{align}
where $\boldsymbol{\theta}^*$ 
are the latent parameters associated with the target input. More specifically, given a test image $\mathbf x^*$, the network performs $T$ inferences with the same dropout rate used during training. We then obtain:
\begin{align}
	p(y^*|\mathbf x^*) = \frac{1}{T} \sum_{t=1}^T p(y^*|\mathbf x^*, \boldsymbol{\theta}^*_t),
	\label{eq::mean}
\end{align}
where $\boldsymbol{\theta}^*_t$ are the weights of the network for the $t^{th}$ inference, and $y^*$ are the averaged predictions. We can evaluate the uncertainty of our range predictions by inspecting the variance of this probability distribution. The final prediction is as follows:
\begin{align}
	\text{$y^{*}_{f}$}=\begin{cases}
		\text{$y^{*}$},  &\text{if} \; \sigma < \lambda y^{*} \\
		\text{0}, &\text{otherwise}
	\end{cases}
	\label{eq::classify}
\end{align}
in which $y^{*}$ is the predicted mean from Eq. \ref{eq::mean}, and $\sigma$ is its standard deviation:
\begin{align}
	\sigma = \sqrt{\dfrac{1}{T} \sum_{i=1}^{T} (y_{i}^{*} - y^{*}) },
\end{align}
where $y_{i}^{*}$ is the value of the $i$th prediction.

The parameter $\lambda$ causes the noise removal threshold to scale linearly with the predicted sensor range, capturing the fact that the noise level worsens with distance from the sensor.
Throughout this paper we choose a value of 0.03 for $\lambda$, as it is found to give the most accurate mapping results, and we choose an inference quantity $T$ of 50 for all experiments. A larger $T$ yields improved results, as the true probability distribution $p(y^*|\mathbf x^*)$ can be better approximated with more predictions.

Since the predictions of this step are between 0 and 1, to obtain the high-res point cloud, we first multiply the range image by the normalization value used in Sec. \ref{sec::data-augmentation}. Then we project the high-res range image back to a high-res point cloud. Note that we do not generate points from the zero-valued pixels in the range image, as they are yielded either from noisy predictions deleted from the range image, occluded objects, or objects outside the sensor's range.




\section{Experiments}
\label{sec:experiments}

\begin{figure}[h]
	\centering
	\includegraphics[width=.80\columnwidth]{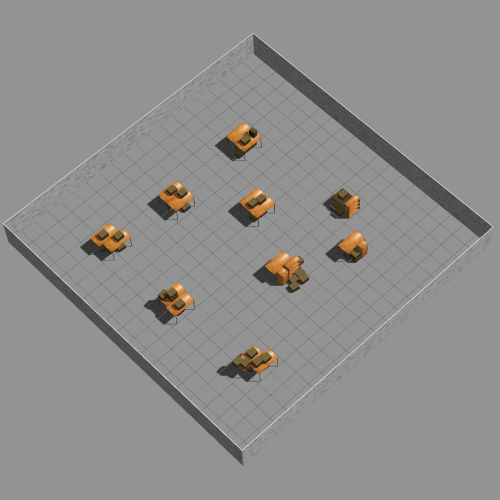}
	\caption{A total number of 25 scans are obtained in this simulated office-like environment for generating Octomaps using different methods.}
	\label{fig::gazebo-environment}
	\vspace{-3mm}
\end{figure}

\begin{figure}[h]
	\centering
	\subfigure[Raw scan]{\includegraphics[width=.48\columnwidth]{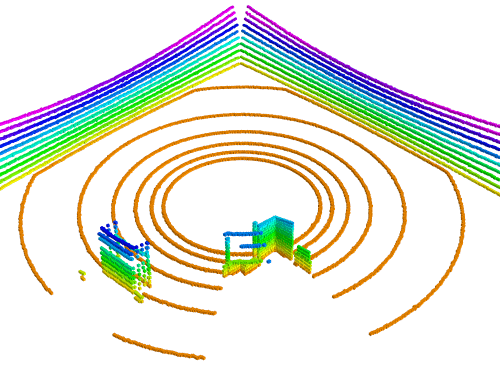}}
	\subfigure[Linear]{\includegraphics[width=.48\columnwidth]{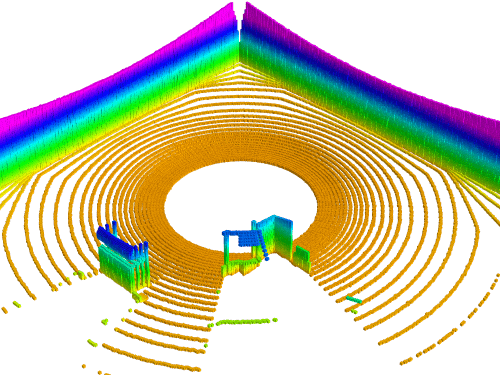}}
	\subfigure[Cubic]{\includegraphics[width=.48\columnwidth]{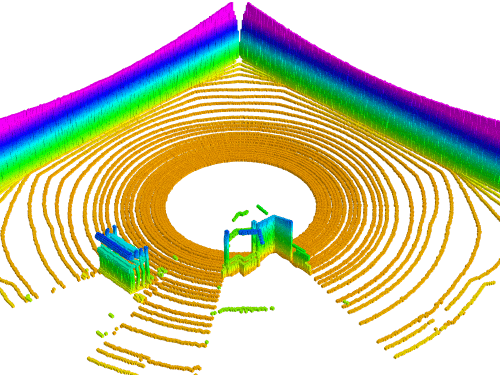}}
	\subfigure[Ours w/o MC-dropout]{\includegraphics[width=.48\columnwidth]{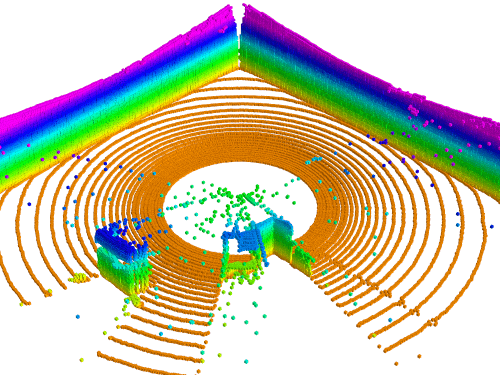}}
	\subfigure[Ours w/ MC-dropout]{\includegraphics[width=.48\columnwidth]{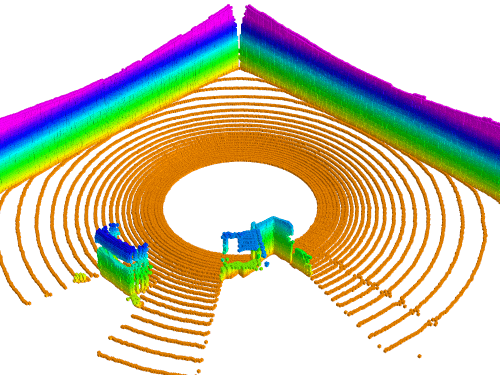}}
	\subfigure[Ground truth scan]{\includegraphics[width=.48\columnwidth]{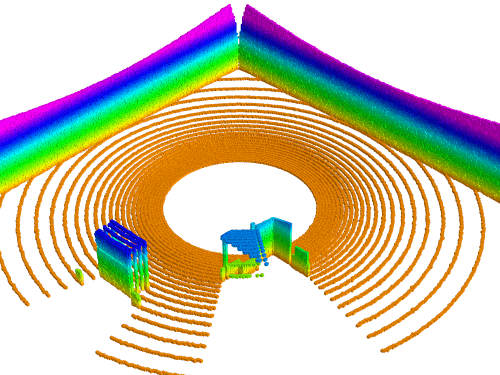}}
	\caption{Lidar scans of an example input (a), predictions by linear and cubic interpolation (b and c), and our methods without and with MC-dropout (d and e) and ground truth (f). As is shown in (d), the inferred point cloud is noisy due to points that have high uncertainty, motivating our use of MC-dropout. Color variation indicates elevation change.}
	\label{fig::gazebo-scan-demo}
	\vspace{-3mm}
\end{figure}

\begin{figure}[h]
	\subfigure[Baseline]{\includegraphics[width=.48\columnwidth]{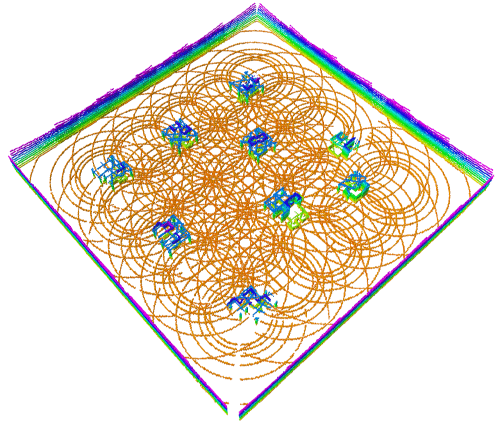}}
	\subfigure[Linear]{\includegraphics[width=.48\columnwidth]{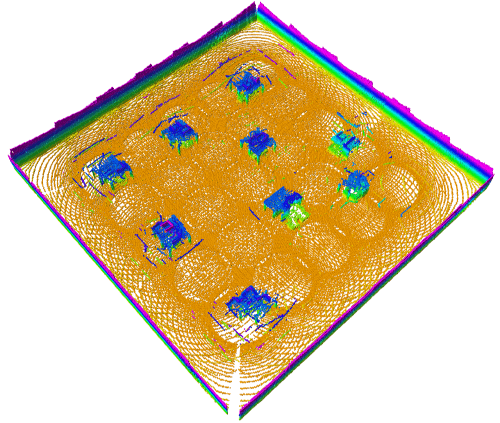}}
	\subfigure[Cubic]{\includegraphics[width=.48\columnwidth]{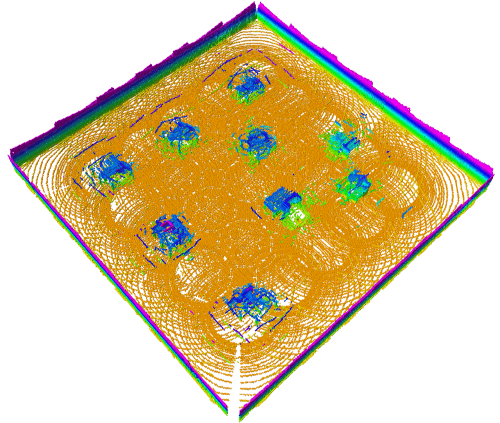}}
	\subfigure[Ours w/o MC-dropout]{\includegraphics[width=.48\columnwidth]{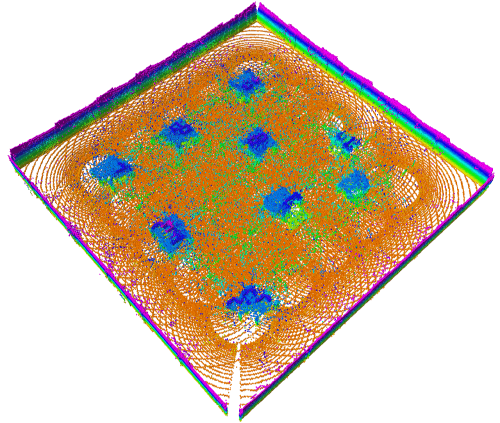}}
	\subfigure[Ours w/ MC-dropout]{\includegraphics[width=.48\columnwidth]{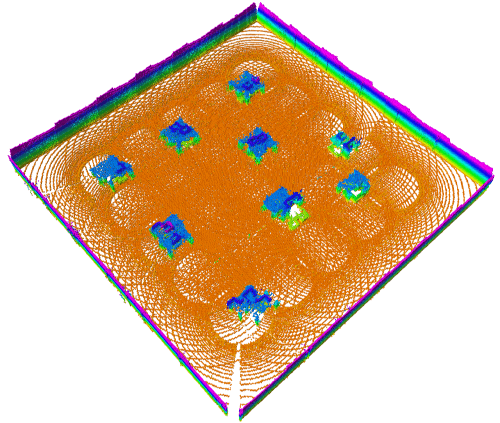}}
	\subfigure[Ground truth]{\includegraphics[width=.48\columnwidth]{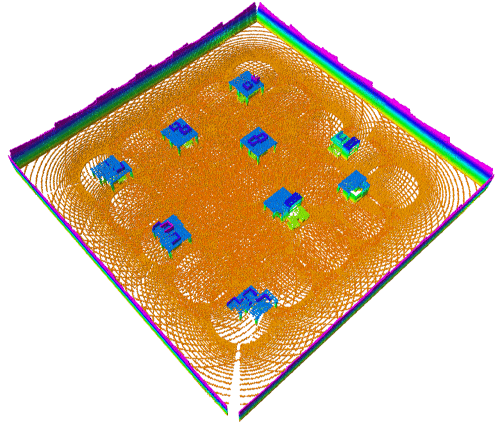}}
	\caption{Full occupancy mapping results generated using the simulated indoor dataset. Color variation indicates elevation change.}
	\label{fig::gazebo-global-demo}
	\vspace{-3mm}
\end{figure}

We now describe a series of experiments to quantitatively and qualitatively analyze the performance of our lidar super-resolution architecture. We perform 4$\times$ upscaling (16 to 64) for all experiments in this section. For more experimental results, please refer to the supplementary Appendix.

For network training, Adam optimizer \cite{adam-optimizer} is used with a learning rate of $10^{-4}$ and decay factor of $10^{-5}$ after each epoch. $\mathcal{L}1$ loss, the sum of absolute differences between the true values and the predicted values associated with range image pixels, is utilized for penalizing the differences between the network output and ground truth, as it achieves high accuracy, fast convergence and improved stability during training. A computer equipped with a Titan RTX GPU was used for training and testing. The training framework was implemented in Keras \cite{keras} using Tensorflow \cite{tensorflow} as a backend in Ubuntu Linux. The software package of the proposed method is publicly available\footnote{\url{https://github.com/RobustFieldAutonomyLab/lidar_super_resolution}}.

Besides benchmarking various methods in 2D image space using $\mathcal{L}1$ loss, we also show that our method is able to produce dense Octomaps \cite{octomap} with high accuracy in 3D Euclidean space. 3D occupancy maps can support a variety of robotics applications, e.g., planning \cite{belief-roadmap-search} and exploration \cite{exploration-1, exploration-2}. However, sparsity in the point cloud of a 3D lidar can leave gaps and inconsistencies in traditional occupancy grid maps, which can be misleading when applied in planning and exploration scenarios. Intuitively, 3D occupancy mapping can benefit greatly from a higher resolution lidar. We use receiver operating characteristic (ROC) curves to benchmark the predictive accuracy (with respect to the binary classification of occupancy) of each method. The ROC curves plot the true positive rate against the false positive rate. We compare all methods to the ground-truth occupancy (0 - free, 0.5 - unknown, 1 - occupied) for all cells in the map. The area under the curve (AUC) is provided for each method for comparison of prediction accuracy. We treat the underlying 64-channel range scan as ground truth, rather than a complete map with all cells filled, because our specific goal is to truthfully compare the range prediction accuracy of each method.

For the simulated experiments described in Sec. \ref{sec::gazebo} and \ref{sec::carla}, we use the exact same neural network to demonstrate that the proposed method is capable of performing accurate prediction for sensors with different mounting positions in different environments. The training data for the neural network is gathered from CARLA Town 02, which features an urban environment, by simulating a 64-channel lidar ``VLP-64" that has a vertical FOV of 30$^{\circ}$. A low-res 16-channel lidar scan is obtained by evenly extracting 16-channel data from the high-res data. The low-res data here is equivalent to the scan obtained from the VLP-16. The training dataset contains 20,000 scans after data augmentation.

Since the real-world Ouster lidar used in Sec. \ref{sec::ouster} has a different FOV (33.2$^{\circ}$), we gather a new training dataset for network training (see Sec. \ref{sec::data-gathering}). Similarly, we simulate a 64-channel lidar, the OS-1-64, in CARLA Town 02 and gather high-res data. The 16-channel data is extracted in the same way as described before. The low-res data here is equivalent to the scan obtained from an OS-1-16 sensor. The training dataset also contains 20,000 scans after data augmentation.

\subsection{Simulated indoor dataset}
\label{sec::gazebo}

We first demonstrate the benefits of applying MC-dropout. We simulate a 64-channel lidar ``VLP-64" and gather 25 high-res scans in an office-like environment in Gazebo. The lidar is assumed to be installed on top of a small unmanned ground vehicle (UGV). The sensor is 0.5m above the ground. As is shown in Fig. \ref{fig::gazebo-environment}, the environment is populated with desks and boxes. The low-res 16-channel testing scans are obtained by evenly extracting 16-channel data from the high-res data. Note that none of these scans are used for network training, nor is the height at which the sensor is mounted. 

\begin{figure*}[h]
	\centering
	\subfigure[Simulated indoor dataset]{\includegraphics[width=.315\textwidth]{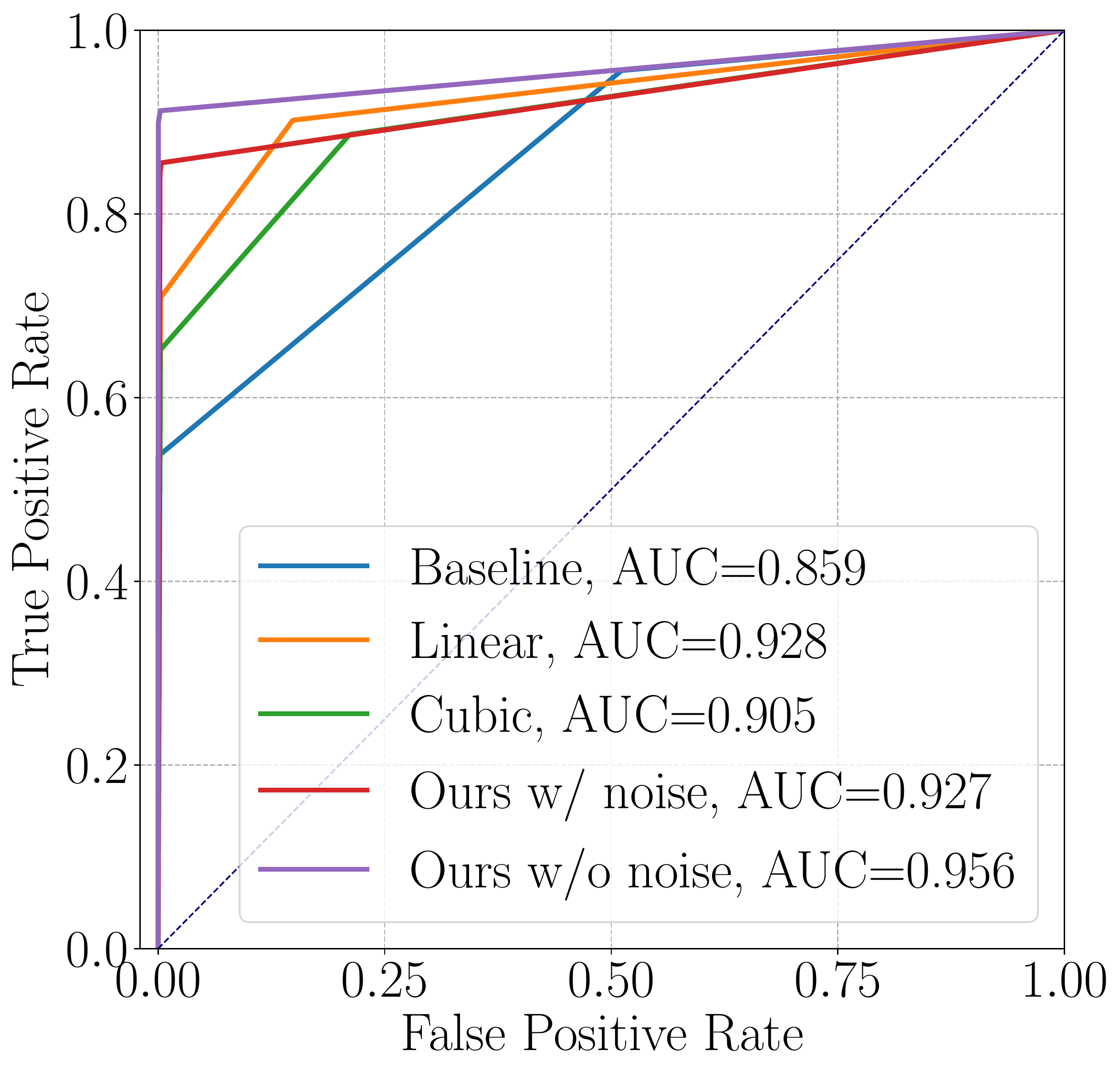}}
	\subfigure[Simulated outdoor dataset]{\includegraphics[width=.3\textwidth]{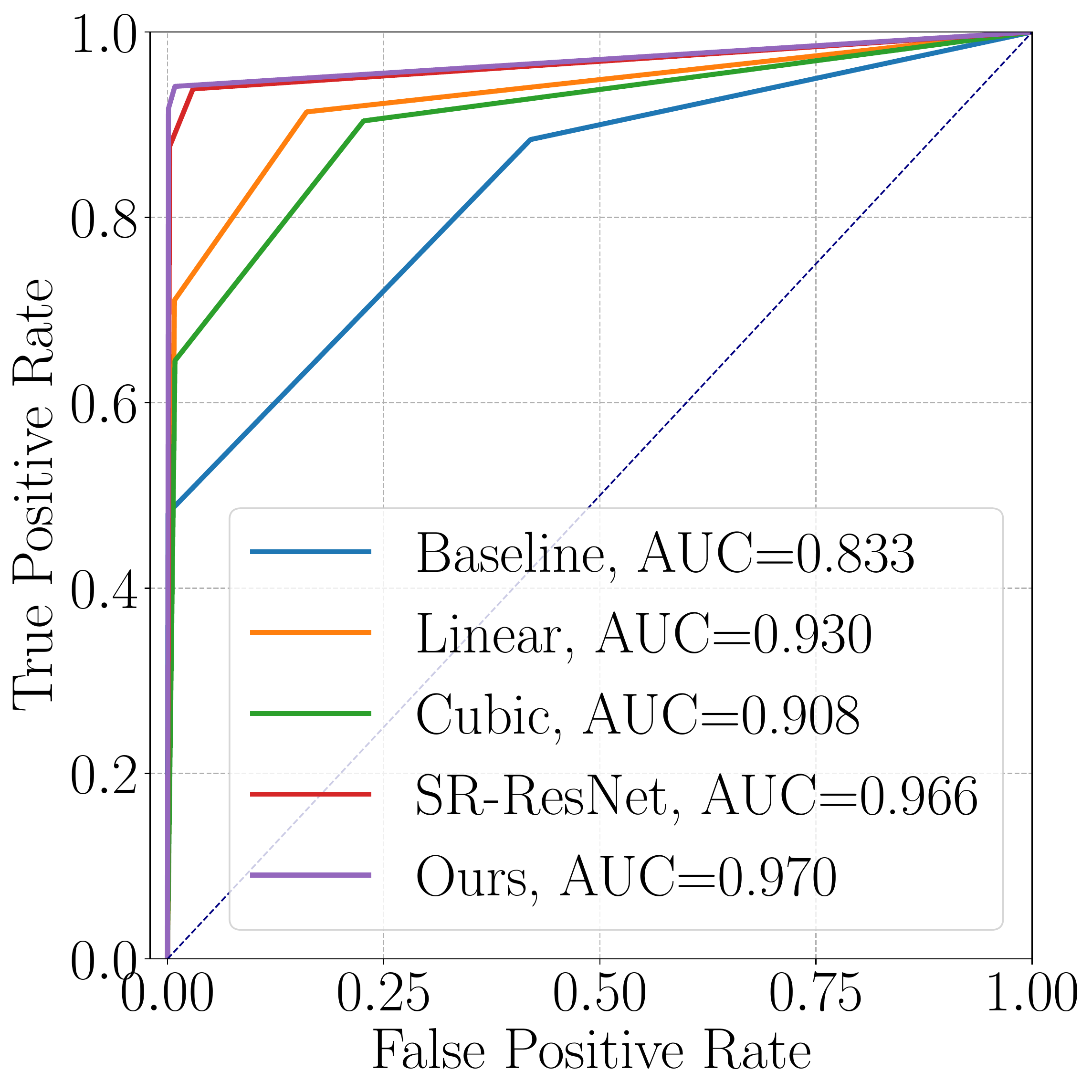}}
	\subfigure[Ouster dataset]{\includegraphics[width=.3\textwidth]{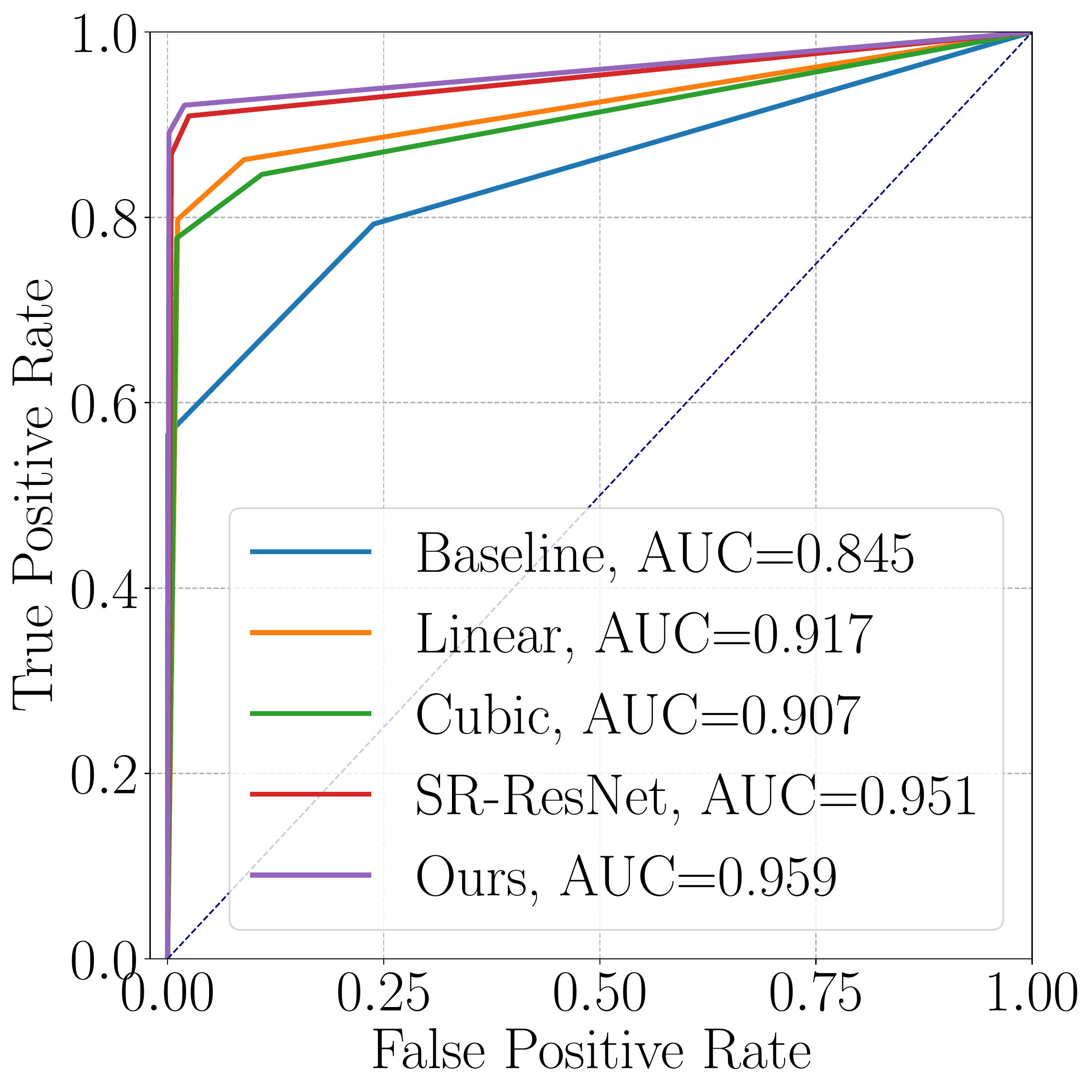}}
	\vspace{-3mm}
	\captionof{figure}{ROC curves and AUC for all competing methods. The results are obtained by comparing the Octomaps of each method with the ground truth Octomap. Though the neural network is trained using the data from a completely different map (CARLA Town 02), our proposed method produces dense Octomaps with the highest AUC among all methods evalutes in all experiments.}
	\label{fig::roc}
\end{figure*}

A representative low-res scan is shown in Fig. \ref{fig::gazebo-scan-demo}(a). Using this scan as input, the predicted high-res scans using the naive linear and cubic interpolation methods are shown in Fig. \ref{fig::gazebo-scan-demo}(b) and (c). The predictions using our method w/ and w/o the application of MC-dropout are shown in Fig. \ref{fig::gazebo-scan-demo}(d) and (e). Without applying MC-dropout, the range prediction is noticeably noisy due to the smoothing effect caused by convolution, hence the scan shown in Fig. \ref{fig::gazebo-scan-demo}(d). After noise removal by applying MC-dropout, the predicted scan shows significantly less noise and resembles the scan of ground truth. The resulting maps are shown in Fig. \ref{fig::gazebo-global-demo}. All the Octomaps have a resolution of 0.05m. We refer to the approach of using low-res lidar scans to produce an Octomap as the $baseline$ approach. The ground truth Octomap is obtained by using the high-res scans.  The map of baseline approach is sparse, as no inference is performed. As is shown in Fig. \ref{fig::gazebo-global-demo}(e), the proposed method is able to produce a dense Octomap that resembles the ground truth Octomap. The AUC and ROC curves of each method are shown in Fig. \ref{fig::roc}(a). The AUC is improved when applying MC-dropout.

We also note that though the network is trained using data from an outdoor environment - CARLA Town 02, our method is capable of producing meaningful and accurate predictions for indoor usage, with a different sensor mounting scheme. This demonstrates that the network is able to learn the complex mapping between low-res input and high-res output while properly maintaining the structure of surrounding objects.
More comparisons of this experiment can be found in the appendix.

\subsection{Simulated outdoor dataset}
\label{sec::carla}

\begin{figure}[h]
	\centering
	\subfigure[]{\includegraphics[width=.45\columnwidth]{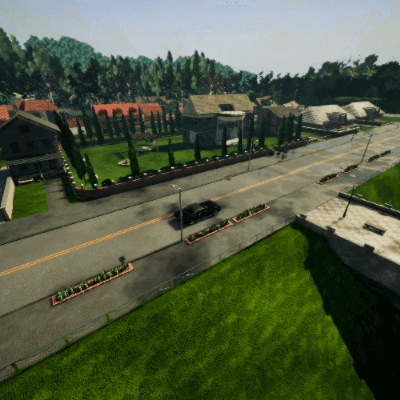}}
	\subfigure[]{\includegraphics[width=.45\columnwidth]{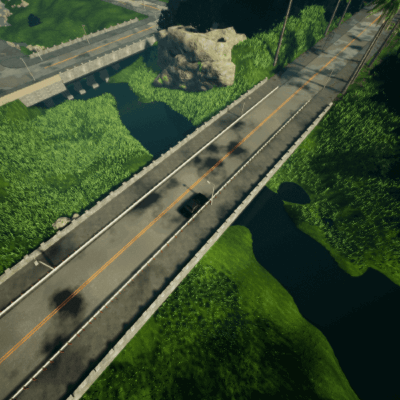}}
	\vspace{-3mm}
	\caption{Two representative scenes from CARLA Town 01. The resulting Octomaps are shown in Fig. \ref{fig::carla-house} and Fig. \ref{fig::carla-bridge}.}
	\label{fig::carla-house-bridge}
	\vspace{-3mm}
\end{figure}

\begin{figure}[h]
	\centering
	\subfigure[Baseline]{\includegraphics[width=.32\columnwidth]{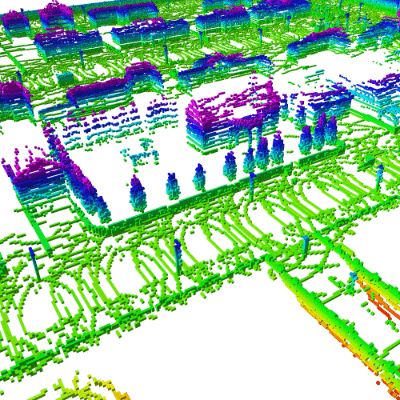}}
	\subfigure[Ground truth]{\includegraphics[width=.32\columnwidth]{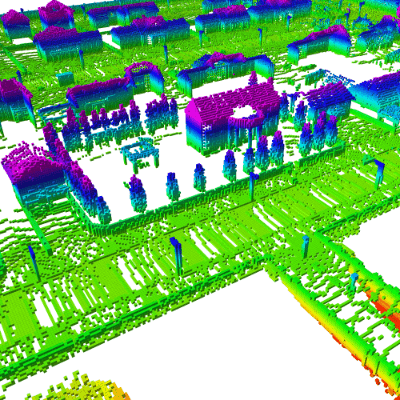}}
	\subfigure[Linear]{\includegraphics[width=.32\columnwidth]{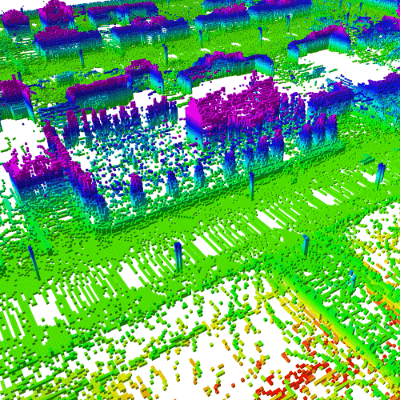}}
	\subfigure[Cubic]{\includegraphics[width=.32\columnwidth]{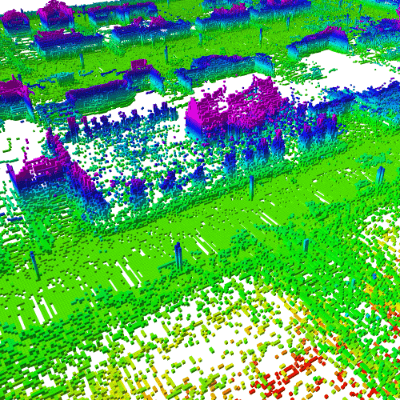}}
	\subfigure[SR-ResNet]{\includegraphics[width=.32\columnwidth]{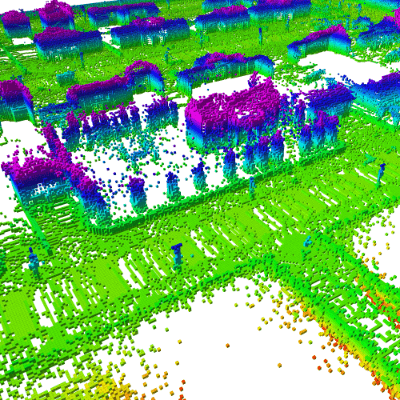}}
	\subfigure[Ours]{\includegraphics[width=.32\columnwidth]{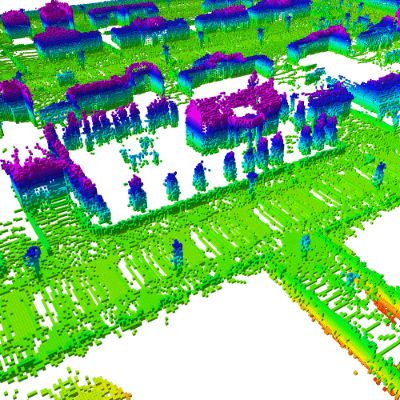}}
	\vspace{-3mm}
	\caption{Occupancy mapping results for scene shown in Fig. \ref{fig::carla-house-bridge}(a). Color variation indicates elevation change.}
	\label{fig::carla-house}
	\vspace{-3mm}
\end{figure}


\begin{figure}[h]
	\centering
	\subfigure[Baseline]{\includegraphics[width=.32\columnwidth]{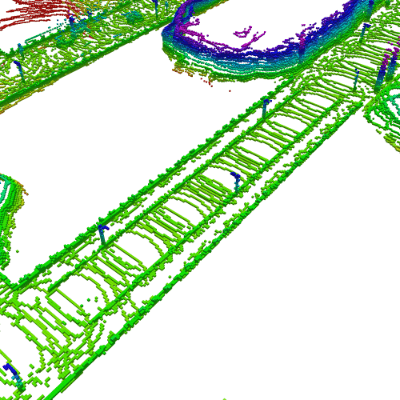}}
	\subfigure[Ground truth]{\includegraphics[width=.32\columnwidth]{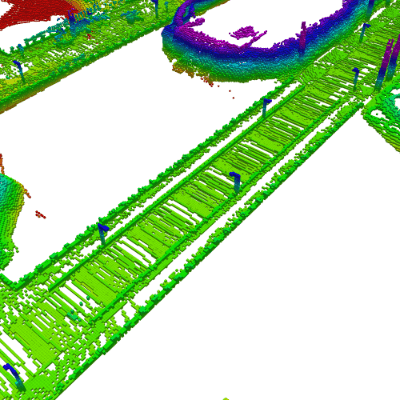}}
	\subfigure[Linear]{\includegraphics[width=.32\columnwidth]{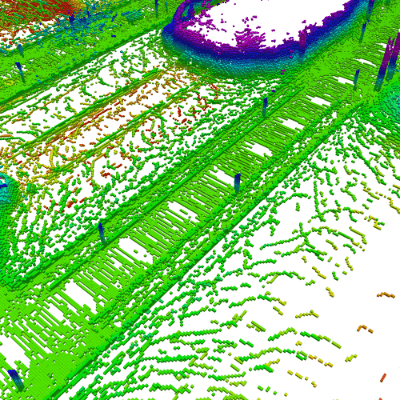}}
	\subfigure[Cubic]{\includegraphics[width=.32\columnwidth]{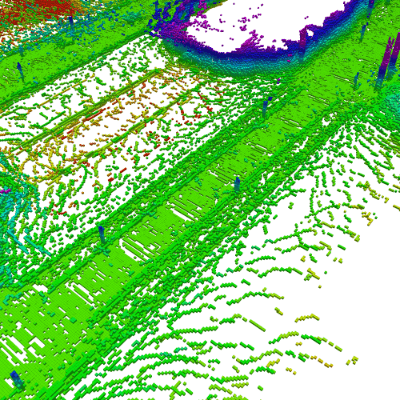}}
	\subfigure[SR-ResNet]{\includegraphics[width=.32\columnwidth]{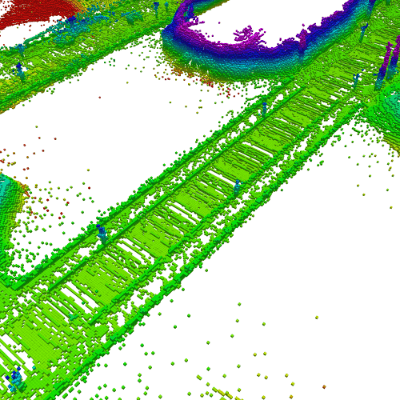}}
	\subfigure[Ours]{\includegraphics[width=.32\columnwidth]{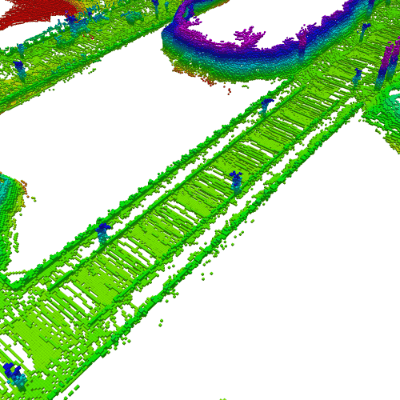}}
	\vspace{-3mm}
	\caption{Occupancy mapping results for scene shown in Fig. \ref{fig::carla-house-bridge}(b). Color variation indicates elevation change. 
	}
	\label{fig::carla-bridge}
	\vspace{-3mm}
\end{figure}

In this experiment, we compare our method with various approaches, which include the standard linear and cubic interpolation techniques and also the state-of-the-art super-resolution approach - SR-ResNet \cite{SR-GAN}, using a simulated large scale outdoor dataset that is gathered in CARLA Town 01. CARLA Town 01 features a suburban environment with roads, trees, houses, and a variety of terrain. The same sensor that is used in \ref{sec::gazebo} is used here. The  ``VLP-64" sensor, which has a height of 1.8m from the ground, is mounted on top of a full-sized passenger vehicle. We drive the vehicle along a trajectory of 3300m and gather a lidar scan every 10m. Thus this dataset contains 330 scans.

The $\mathcal{L}1$ loss of each method is shown in Table \ref{table::benchmark}. The deep learning approaches outperform the traditional interpolation approaches by a large margin. For fair comparison, we also apply MC-dropout on SR-ResNet by adding a dropout layer to the end of each residual block for noise removal. The losses of SR-ResNet and our method are very close. However, the amount of noise removed per scan from SR-ResNet is much larger than our method. Though we can adjust $\lambda$ in Eq. \ref{eq::classify} to retain more points, the mapping accuracy deteriorates greatly as more noisy points are introduced. We can also decrease the value of $\lambda$ for SR-ResNet to filter out more noise. The mapping accuracy then also deteriorates, as more areas in the map become unknown.

The Octomaps of the competing methods using a low-res scan as input are shown in Fig. \ref{fig::carla-house} and  \ref{fig::carla-bridge}. The baseline approach naturally yields the most sparse map. Though offering better coverage, the Octomaps of the linear and cubic methods are very noisy due to range interpolation between different objects. 
The deep learning-enabled approaches, SR-ResNet and our proposed method, outperform the interpolation-based approaches by offering true representation of the environment.
Though SR-ResNet outperforms linear and cubic interpolation methods in 2D image space by yielding smaller $\mathcal{L}1$ loss, its predictions, when shown in 3D Euclidean space, still contain a great deal of noise at object boundaries. Our proposed approach introduces much less noise into the map. As a result, our method produces a map that is easier to interpret visually, and which also achieves the highest AUC among all methods. The AUC and ROC curves for each method using 330 scans are shown in Fig. \ref{fig::roc}(b). 

\begin{table}[h]
	\centering
	\caption{Quantitative results for the experiments discussed in Sec. \ref{sec::carla} and \ref{sec::ouster}.}
	\label{table::benchmark}
	\begin{tabular}{@{}cccc@{}}
		\toprule[1pt]
		Dataset & Method & $\mathcal{L}1$ Loss & \begin{tabular}[c]{@{}c@{}}Removed\\ points (\%)\end{tabular} \\
		\midrule[1pt]
		\multirow{4}{*}{\begin{tabular}[c]{@{}c@{}}CARLA\\ Town 01\end{tabular}} & Linear & 0.0184 & N/A \\
		& Cubic & 0.0303 & N/A \\
		& SR-ResNet & 0.0089 & 12.37 \\
		& Ours & 0.0087 & 4.13 \\
		\midrule[1pt]
		\multirow{4}{*}{Ouster} & Linear & 0.0324 & N/A \\
		& Cubic & 0.0467 & N/A \\
		& SR-ResNet & 0.0211 & 17.70 \\
		& Ours & 0.0214 & 8.37 \\
		\bottomrule[1pt]
	\end{tabular}	
\vspace{-5mm}
\end{table}

\subsection{Real-world outdoor dataset}
\label{sec::ouster}

\begin{figure*}[t]
	\centering
	\subfigure[Scene 1]{\includegraphics[width=.138\textwidth]{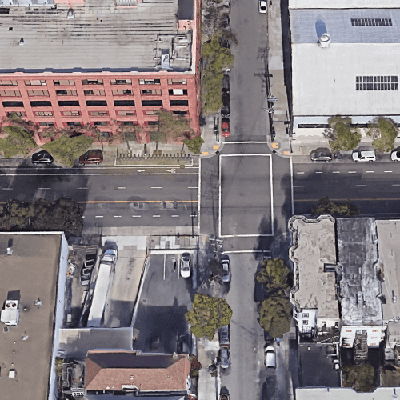}}
	\subfigure[Baseline]{\includegraphics[width=.138\textwidth]{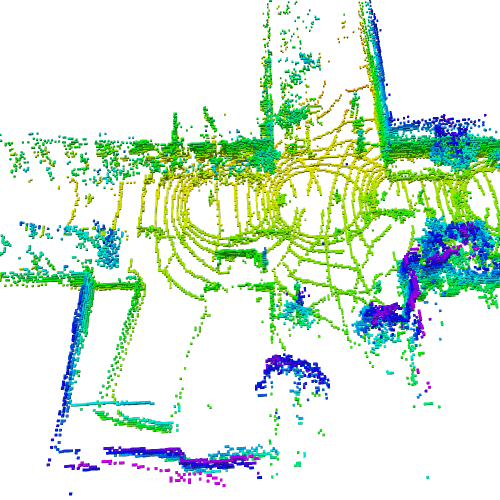}}
	\subfigure[Linear]{\includegraphics[width=.138\textwidth]{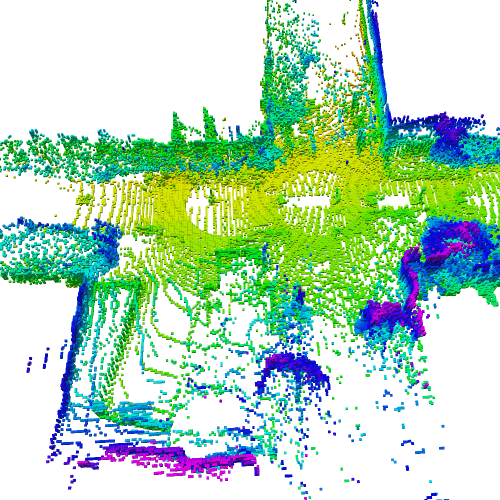}}
	\subfigure[Cubic]{\includegraphics[width=.138\textwidth]{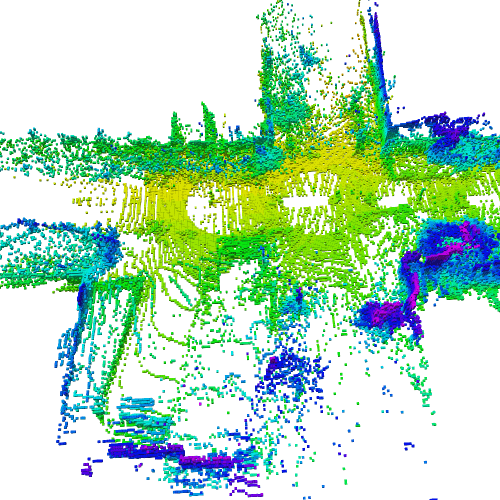}}
	\subfigure[SR-ResNet]{\includegraphics[width=.138\textwidth]{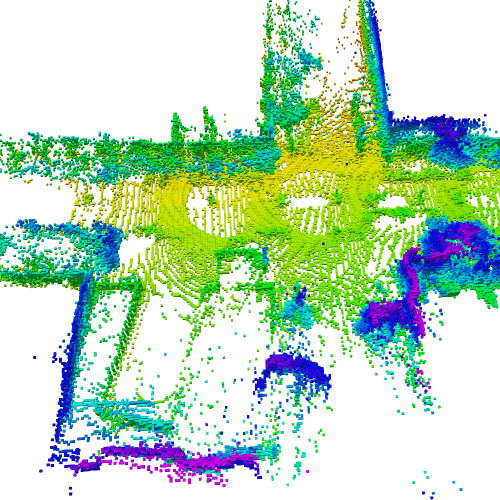}}
	\subfigure[Ours]{\includegraphics[width=.138\textwidth]{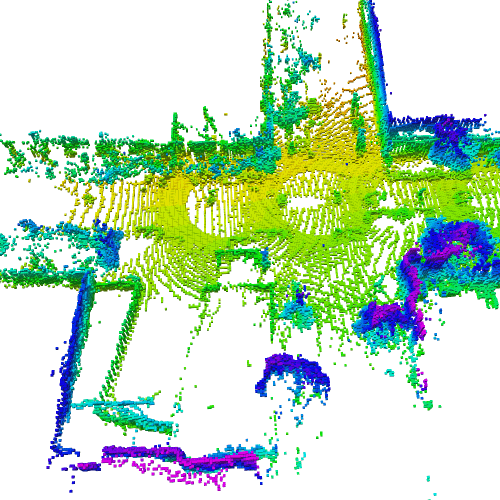}}
	\subfigure[Ground truth]{\includegraphics[width=.138\textwidth]{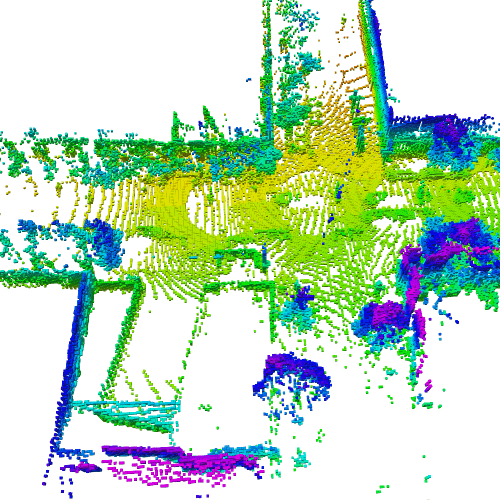}}
	\subfigure[Scene 2]{\includegraphics[width=.138\textwidth]{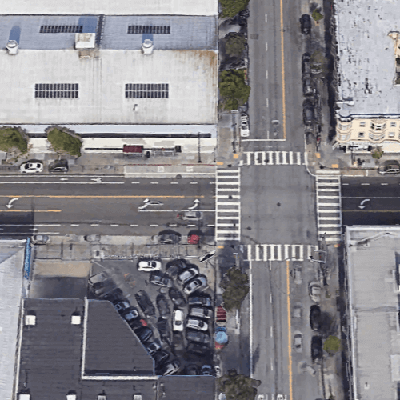}}
	\subfigure[Baseline]{\includegraphics[width=.138\textwidth]{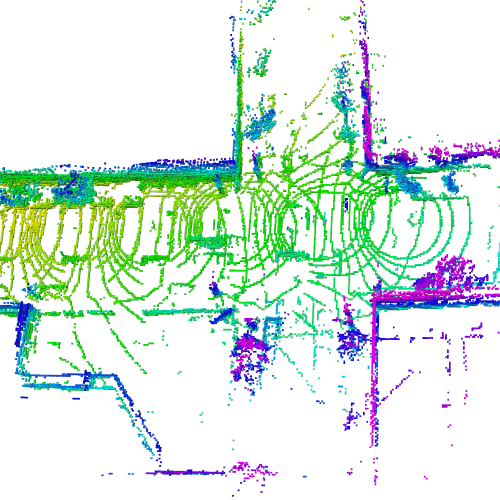}}
	\subfigure[Linear]{\includegraphics[width=.138\textwidth]{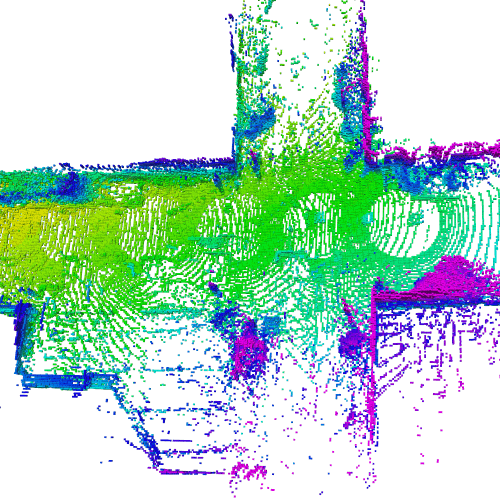}}
	\subfigure[Cubic]{\includegraphics[width=.138\textwidth]{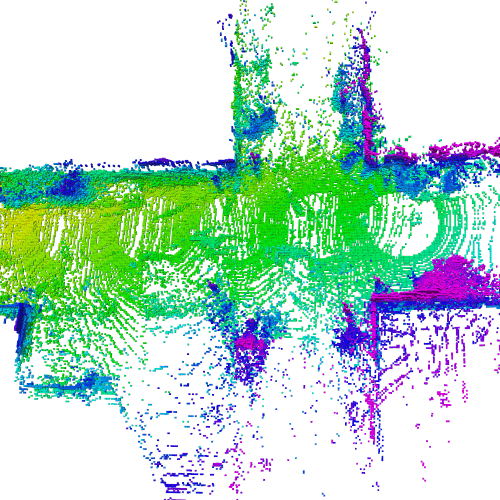}}
	\subfigure[SR-ResNet]{\includegraphics[width=.138\textwidth]{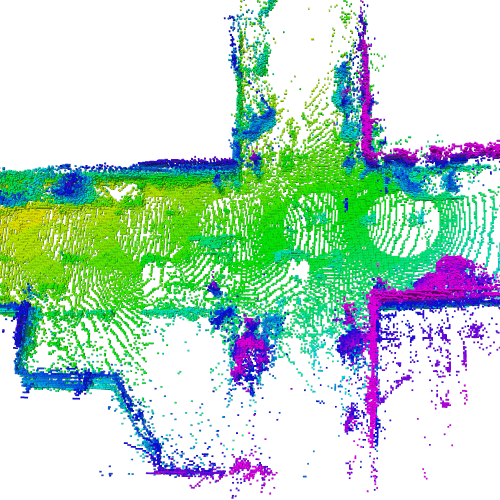}}
	\subfigure[Ground truth]{\includegraphics[width=.138\textwidth]{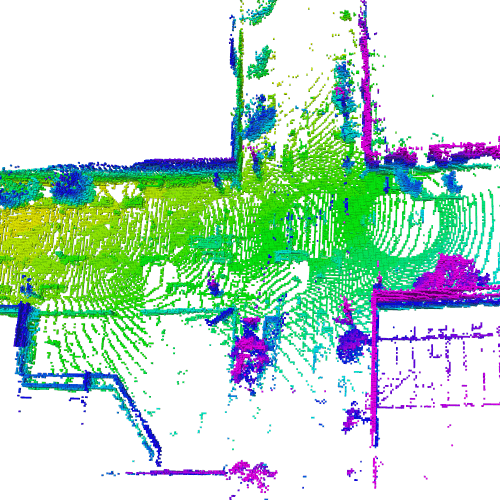}}
	\subfigure[Ours]{\includegraphics[width=.138\textwidth]{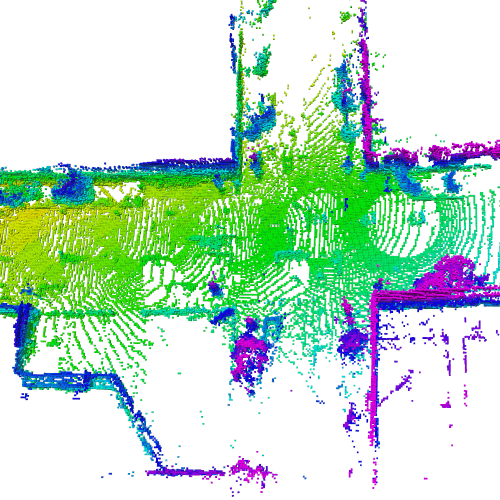}}
	\vspace{-3mm}
	\caption{Occupancy mapping results using the Ouster dataset. Color variation indicates elevation change.}
	\label{fig::ouster-local}
	\vspace{-3mm}
\end{figure*}

In this experiment, we evaluate the proposed method over one publicly available driving dataset, which we refer to as $Ouster$\footnote{\url{https://git.io/fhbBt}}. The Ouster dataset, which consists of 8825 scans over a span of 15 minutes of driving, is captured in San Francisco, CA using an Ouster OS-1-64 3D lidar. This 64-channel sensor naturally gives us the ground truth for validation, as we only need to extract a few channels of data for generating low-res range image inputs. 
Again, the networks evaluated here are purely trained using simulated dataset gathered from CARLA Town 02.
As is shown in Table \ref{table::benchmark}, both SR-ResNet and our method achieve similar $\mathcal{L}1$ loss, which is evaluated over 8825 scans. 
We note that the $\mathcal{L}1$ loss of SR-ResNet is slightly smaller than that of our proposed method. This is because we compute the $\mathcal{L}1$ loss using all the predicted ranges without applying Equation \ref{eq::classify}, to ensure a fair comparison.
However, the percentage of removed points of our approach is much less when compared with the results of SR-ResNet.
This is because the predictions of SR-ResNet are more influenced by the smoothing effects discussed in Sec. \ref{sec::noise-removal}.
In other words, the predictions of our approach are of lower variance.

We use 15 scans from this dataset to obtain real-world low-res and high-res lidar scans, which are then used to obtain Octomaps, in the same way that is described in our previous experiments. The scans are registered using LeGO-LOAM \cite{lego-loam}. The mapping results at two intersections are shown in Fig. \ref{fig::ouster-local}. All the Octomaps have a resolution of 0.3m. The AUC and ROC curves for each method using these 15 scans are shown in Fig. \ref{fig::roc}(c). Again, our proposed approach outperforms all methods by achieving the highest AUC. For the mapping visualization using all 15 scans, please refer to the appendix. A visualization of the inference performed throughout the dataset can be found in our video attachment\footnote{\url{https://youtu.be/rNVTpkz2ggY}}.


\section{Conclusions and Discussion}

We have proposed a lidar super-resolution method that produces high resolution point clouds with high accuracy. Our method transforms the problem from 3D Euclidean space to an image super-resolution problem in 2D image space, and deep learning is utilized to enhance the resolution of a range image, which is projected back into a point cloud. We train our neural network using computer-generated data, which affords the flexibility to consider a wide range of operational scenarios. We further improve the inference accuracy by applying MC-dropout. The proposed method is evaluated on a series of datasets, and the results show that our method can produce realistic high resolution maps with high accuracy. In particular, we evaluate the super-resolution framework through a study of its utility for \textit{occupancy mapping}, since this is a widely useful perceptual end-product that robots may use to support planning, exploration, inspection, and other activities. In addition to the appealing generalizability of up-scaling at the front end, by predicting the measurements of a higher-resolution sensor, our approach also achieves superior accuracy in the end-stage maps produced, as compared with both deep learning methods and simpler interpolation methods. 

Future work may involve using a generative adversarial network to further improve the inference quality. We conducted tests using SR-GAN, which is proposed in \cite{SR-GAN}. However, we did not obtain additional benefits from using this network for our tests. The discriminator is not able to distinguish generated high-res range images from real high-res range images with an accuracy of more than 40\%. We suspect the low accuracy is caused by the lack of texture details in range images, as SR-GAN is proposed for photo-realistic images. We also encounter noisy predictions on irregular objects, such as trees and bushes. One cause of this problem is that the simulated environment is relatively simply structured. Though vegetation appear in simulation, they are only represented by simple geometries. Thus the training data gathered in the simulation possesses significantly less noise when compared with data from real-world environments. Another potential direction for future work may involve training with a combination of real and synthetic data.


\clearpage


\appendix

\section{Supplements for Sec. \ref{sec::gazebo}}

\begin{figure}[t]
	\centering
	\includegraphics[width=.48\columnwidth]{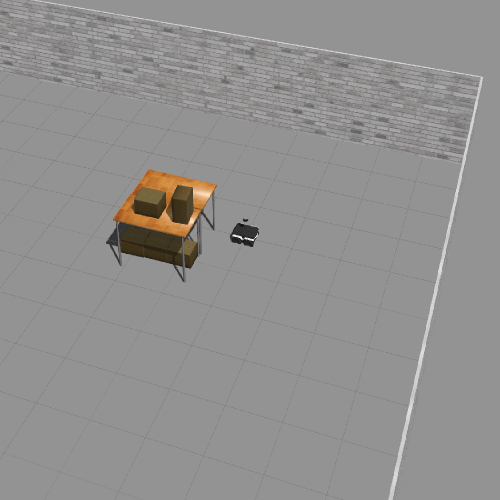}
	\vspace{-2mm}
	\caption{A local region of the environment shown in Fig. \ref{fig::gazebo-environment}.}
	\label{fig::gazebo-environment-local}
	\vspace{-3mm}
\end{figure}

\begin{figure}[t]
	\centering
	\subfigure[Baseline]{\includegraphics[width=.32\columnwidth]{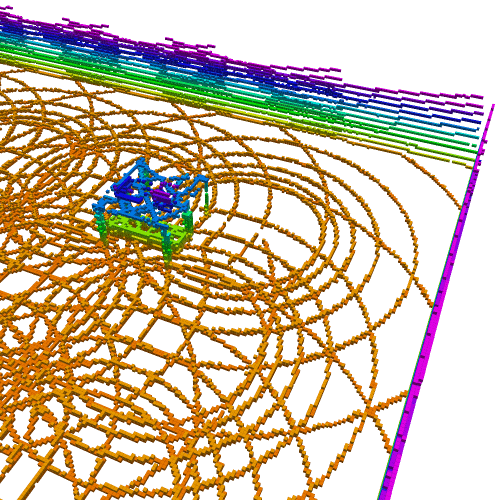}}
	\subfigure[Linear]{\includegraphics[width=.32\columnwidth]{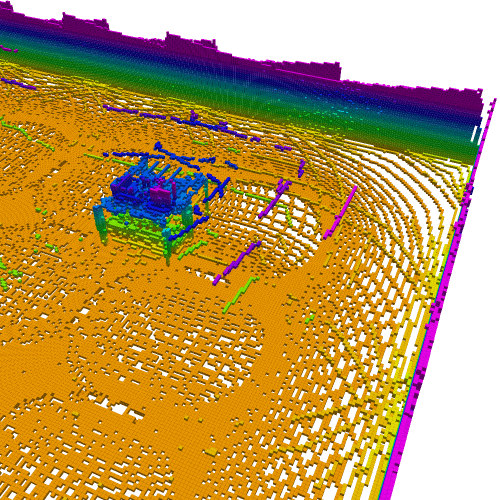}}
	\subfigure[Cubic]{\includegraphics[width=.32\columnwidth]{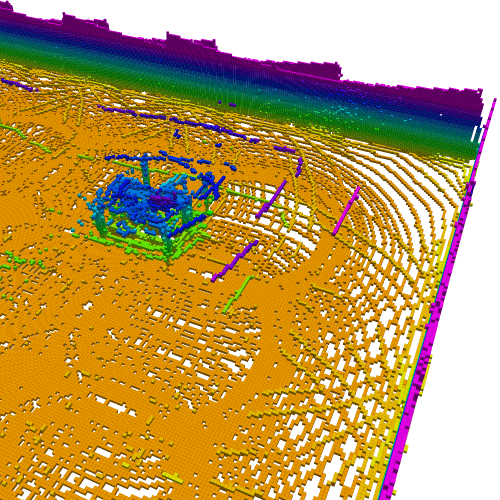}}
	\subfigure[Ours w/o MC-dropout]{\includegraphics[width=.32\columnwidth]{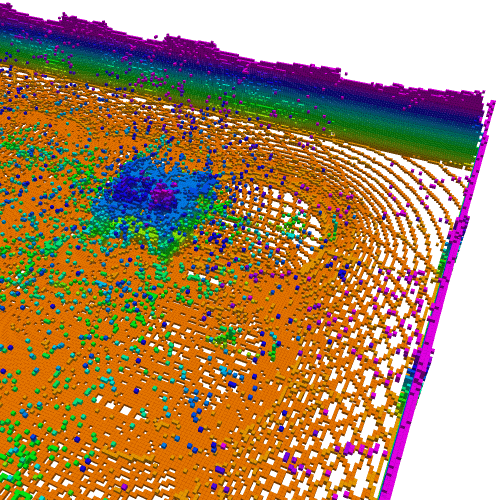}}
	\subfigure[Ours w/ MC-dropout]{\includegraphics[width=.32\columnwidth]{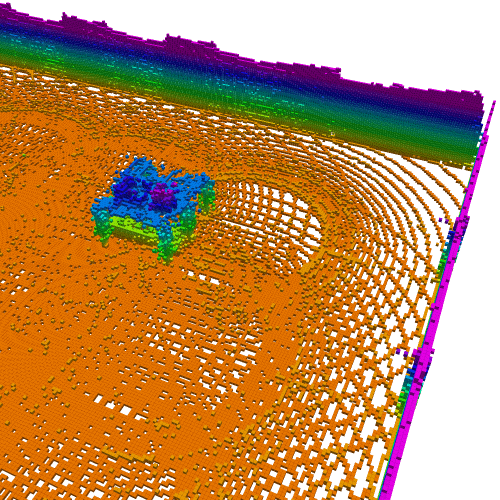}}
	\subfigure[Ground truth]{\includegraphics[width=.32\columnwidth]{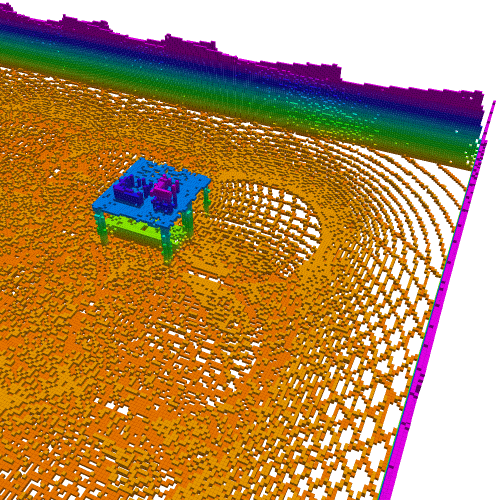}}
	\vspace{-2mm}
	\caption{Mapping results of the area shown in Fig. \ref{fig::gazebo-environment-local} using several competing methods. Color variation indicates elevation change. }
	\label{fig::gazebo-corner-octomap-multi}
	\vspace{-3mm}
\end{figure}

We show detailed Octomaps produced by several methods introduced in Sec. \ref{sec::gazebo}. The scene shown in Fig. \ref{fig::gazebo-environment-local} is located at the top corner of Fig. \ref{fig::gazebo-environment}.
A lidar mounted on top of a small unmanned ground vehicle (UGV), located in the center of the image, is used for capturing the data. The Octomaps for the entire environment are shown in Fig. \ref{fig::gazebo-global-demo}. The resulting Octomaps of this region are shown in Fig. \ref{fig::gazebo-corner-octomap-multi}.
Note that, when using linear or cubic interpolation, the ``ring" structure that covers the ground at the top-right corner differs greatly from the structure of (f).  This is because these methods are interpolating over Euclidean space rather than the sensor's field of view. Naive interpolation methods are not able to retain the real range measurements characterizing the output of a real lidar. As is visible in (b) and (c), these methods also introduce erroneous range measurements by interpolating among the returns from distinctly different objects (such as the floor and above-ground objects). Our proposed deep learning method does not encounter this problem.

\section{Supplements for Sec. \ref{sec::ouster}}


\begin{figure}[t]
	\centering
	\subfigure[Input - 1]{\includegraphics[width=.32\columnwidth]{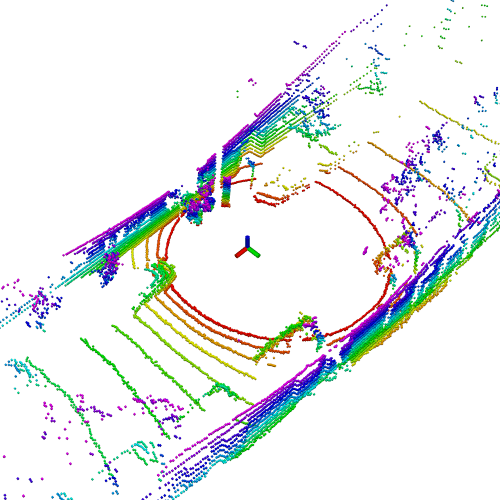}}
	\subfigure[Predicted - 1]{\includegraphics[width=.32\columnwidth]{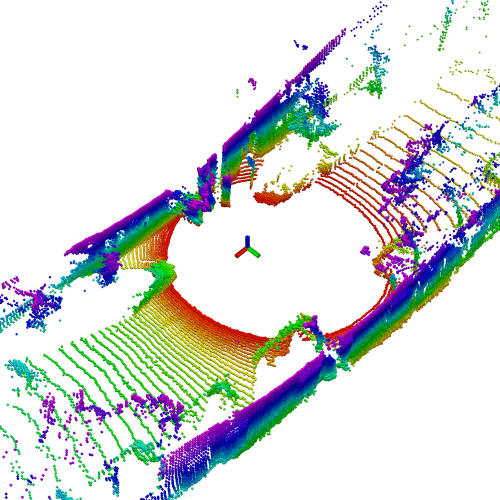}}
	\subfigure[Ground truth - 1]{\includegraphics[width=.32\columnwidth]{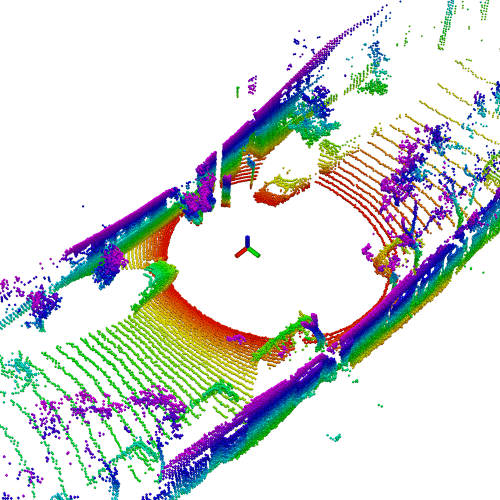}}
	\subfigure[Input - 2]{\includegraphics[width=.32\columnwidth]{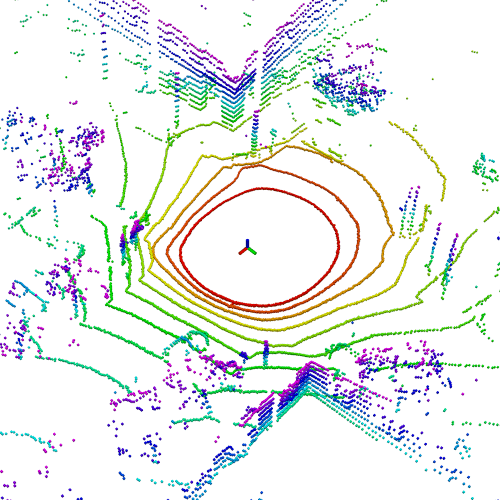}}
	\subfigure[Predicted - 2]{\includegraphics[width=.32\columnwidth]{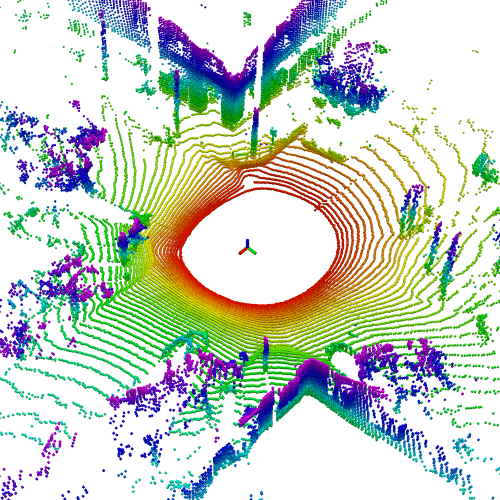}}
	\subfigure[Ground truth - 2]{\includegraphics[width=.32\columnwidth]{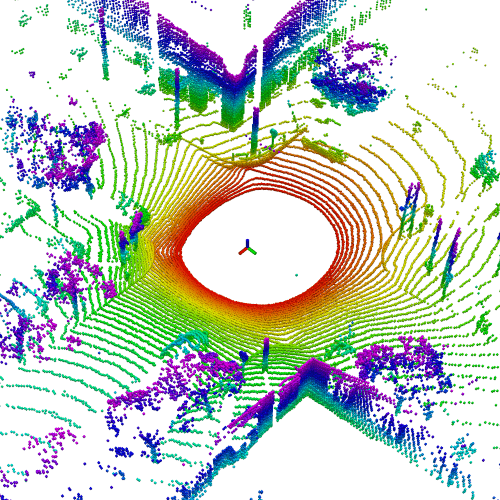}}
	\subfigure[Input - 3]{\includegraphics[width=.32\columnwidth]{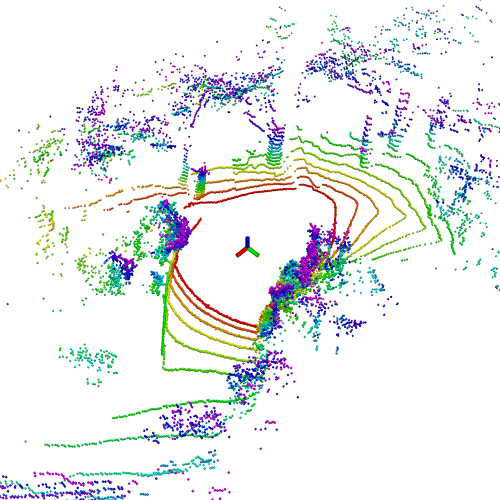}}
	\subfigure[Predicted - 3]{\includegraphics[width=.32\columnwidth]{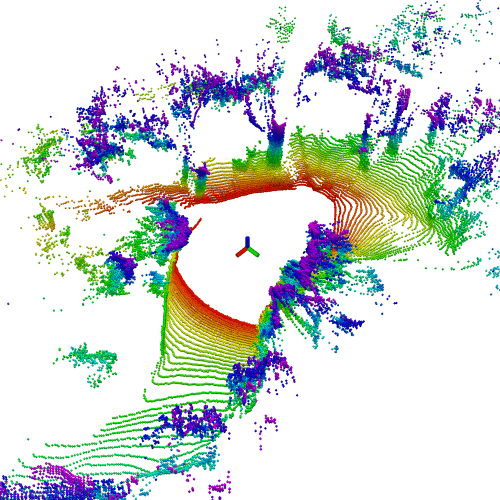}}
	\subfigure[Ground truth - 3]{\includegraphics[width=.32\columnwidth]{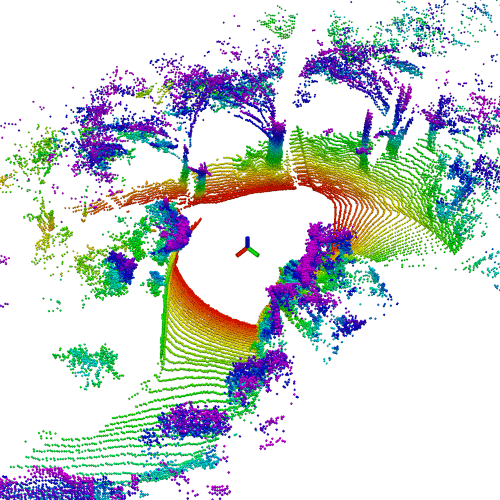}}
	\caption{Visualization of several representative point clouds from the Ouster dataset. The low density point clouds before inference are shown in (a), (d) and (g). The inferred high-res point clouds (4$\times$ upscaling) are shown in (b), (e) and (h). The ground truth point cloud captured by the lidar is shown in (c), (f) and (i). Color variation indicates lidar ``ring" index. }
	\label{fig::ouster-point-cloud-demo}
	\vspace{-3mm}
\end{figure}

\begin{figure}[t]
	\centering
	\subfigure[Range images for scene 1 shown in Fig. \ref{fig::ouster-point-cloud-demo}]{\includegraphics[width=.98\columnwidth]{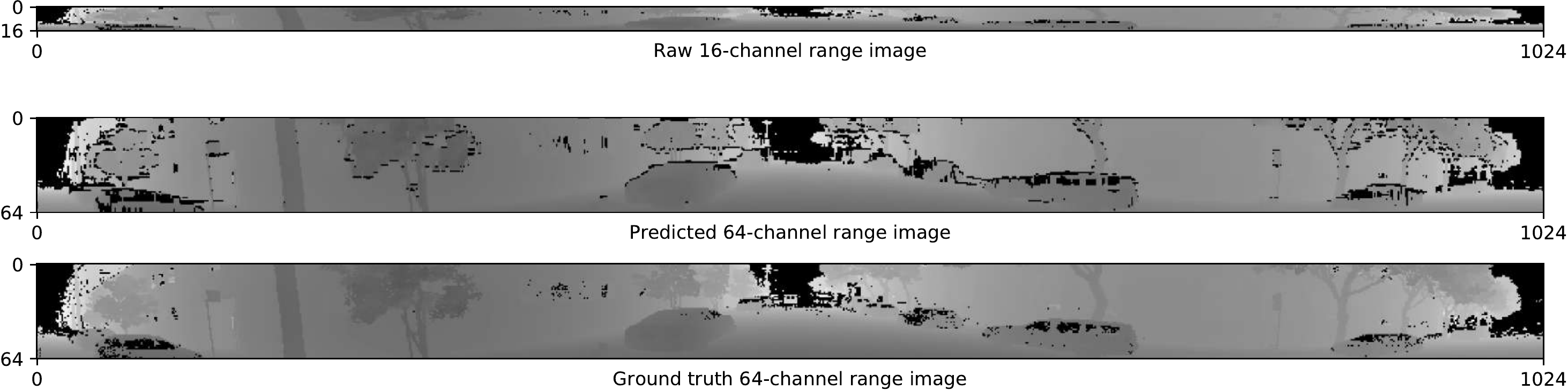}}
	\subfigure[Range images for scene 2 shown in Fig. \ref{fig::ouster-point-cloud-demo}]{\includegraphics[width=.98\columnwidth]{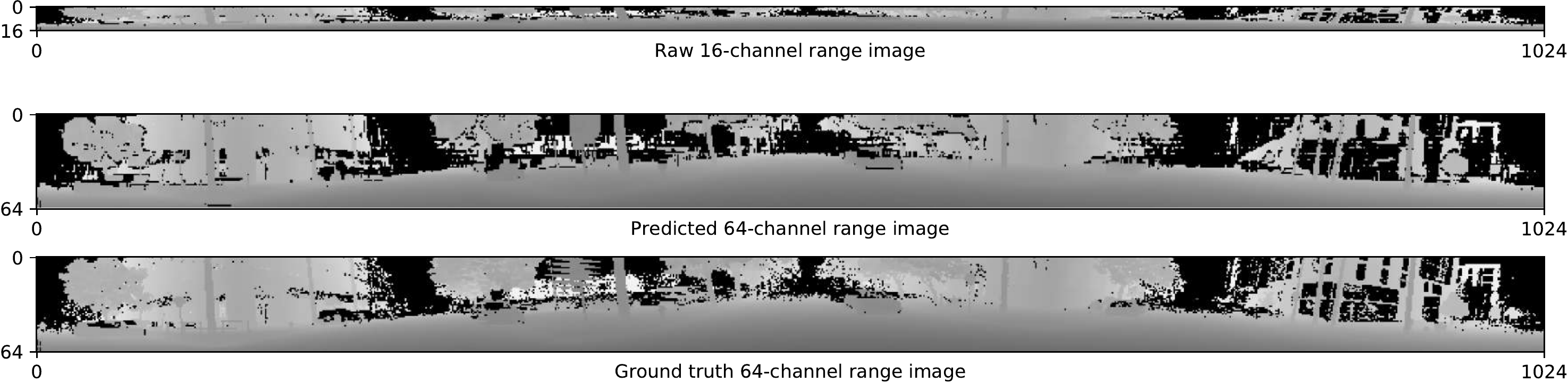}}
	\subfigure[Range images for scene 3 shown in Fig. \ref{fig::ouster-point-cloud-demo}]{\includegraphics[width=.98\columnwidth]{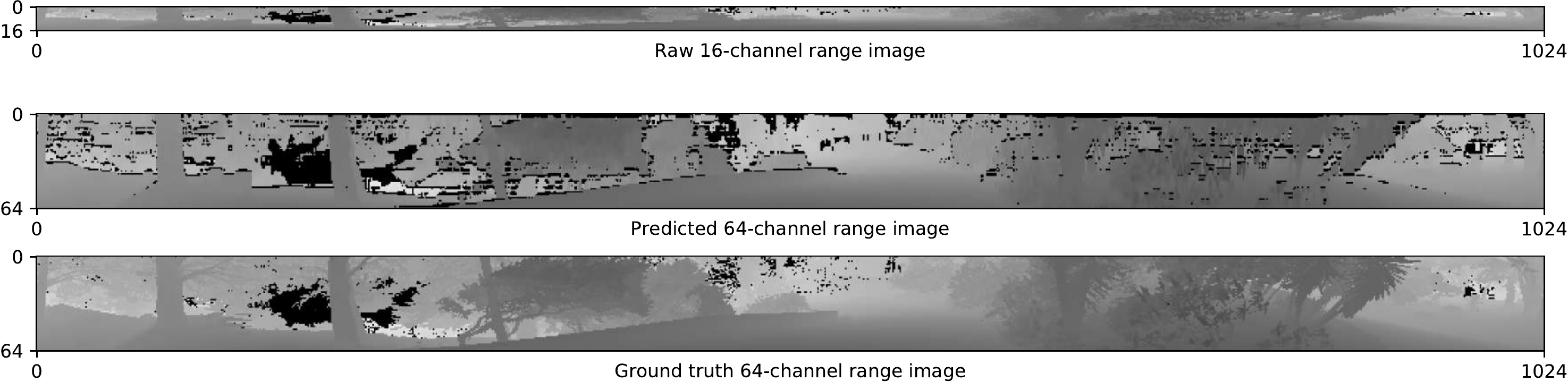}}
	\vspace{-2mm}
	\caption{Range images of the projected point clouds shown in Fig. \ref{fig::ouster-point-cloud-demo}. }
	\label{fig::ouster-range-image-demo}
	\vspace{-3mm}
\end{figure}

\begin{table*}[t]
	\centering
	\caption{Quantitative results for testing using Ouster dataset with different upscailing factors}
	\label{tab:ouster-various-upscaling}
	\begin{tabular}{@{}cccccc@{}}
		\toprule[1pt]
		\begin{tabular}[c]{@{}c@{}}Upscaling\\ factor\end{tabular} & Method & $\mathcal{L}1$ Loss & \begin{tabular}[c]{@{}c@{}}Removed\\ pixels (\%)\end{tabular} & \begin{tabular}[c]{@{}c@{}}Training time\\ per image (ms)\end{tabular} & \begin{tabular}[c]{@{}c@{}}Testing time\\ per image(ms)\end{tabular} \\
		\midrule[1pt]
		\multirow{4}{*}{8x (8 to 64)} & Linear & 0.0455 & N/A & N/A & N/A \\
		& Cubic & 0.0595 & N/A & N/A & N/A \\
		& SR-ResNet & 0.0320 & 30.70 & 21 & 7 \\
		& Ours & 0.0318 & 15.71 & 18 & 6 \\
		\midrule[1pt]
		\multirow{4}{*}{4x (16 to 64)} & Linear & 0.0324 & N/A & N/A & N/A \\
		& Cubic & 0.0467 & N/A & N/A & N/A \\
		& SR-ResNet & 0.0211 & 17.70 & 25 & 7 \\
		& Ours & 0.0214 & 8.37 & 18 & 6 \\
		\midrule[1pt]
		\multirow{4}{*}{2x (32 to 64)} & Linear & 0.0213 & N/A & N/A & N/A \\
		& Cubic & 0.0346 & N/A & N/A & N/A \\
		& SR-ResNet & 0.0118 & 8.92 & 29 & 9 \\
		& Ours & 0.0117 & 2.38 & 17 & 6 \\
		\bottomrule[1pt]
	\end{tabular}
\end{table*}

We give further detailed benchmarking results by comparing four metrics using different upscailing factors:
(1) $\mathcal{L}1$ loss, which is the inference loss tested on the Ouster dataset using the network trained with simulated dataset from CARLA Town 02.
(2) Removed pixels, which indicates the mean percentage of pixels deleted from the range images of each dataset.
(3) Training time, which is the processing time for each range image during training on the simulated dataset from CARLA Town 02.
(4) Testing time, which is the processing time for each range image during testing on the Ouster dataset. 

As is shown in Table \ref{tab:ouster-various-upscaling}, We have observed that the prediction accuracy degrades when less resolution is provided. Both SR-ResNet and our method achieve similar loss over different upscaling factors. However, the percentage of removed pixels of our method is much less when compared with the results of SR-ResNet. In other words, the predictions of our approach are of lower variance. Additionally, the proposed approach requires up to 45\% less time to train the neural network. Note that all the results in the table are evaluated and averaged over 8825 scans.

We also note that the testing time of our proposed framework per image, which does not exceed 10ms for any of the upscaling factors considered, is compatible with real-time performance over the lidars considered in this paper. Scanning at 10Hz, these lidars would require the testing time not to exceed 100ms, but our framework is well within this real-time performance envelope using the modest arrangement of hardware described in Section \ref{sec:experiments}, which is compatible with embedded systems.

We also show qualitative results of performing 4$\times$ upscaling inference using our method on the Ouster dataset in Fig. \ref{fig::ouster-point-cloud-demo}. The three representative point clouds are captured from a narrow street, an open intersection, and a slope surrounded by vegetation, respectively. We can observe that objects such as buildings, roads and pillars are inferred well.

The corresponding range images of the projected point clouds are shown in Fig. \ref{fig::ouster-range-image-demo}. Black color indicates a range value of zero, for which no points are added to the point cloud. Since performing convolutional operations on a range image will unavoidably cause smoothing effects on sharp and discontinuous object boundaries, we apply MC-dropout to identify these erroneous predictions. Accordingly, the range predictions with high variance are removed.

\begin{figure}[h!]
	\centering
	\subfigure[Scene 1]{\includegraphics[width=.49\columnwidth]{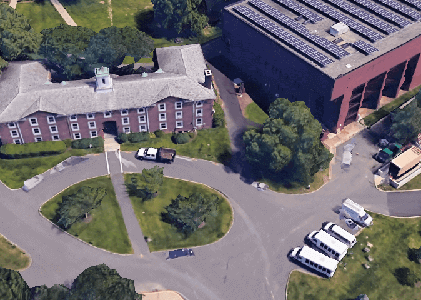}}
	\subfigure[Scene 2]{\includegraphics[width=.49\columnwidth]{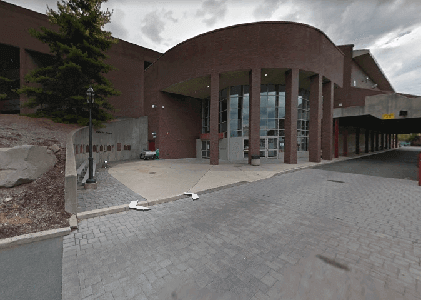}}
	\subfigure[Raw VLP-16 scan 1]{\includegraphics[width=.49\columnwidth]{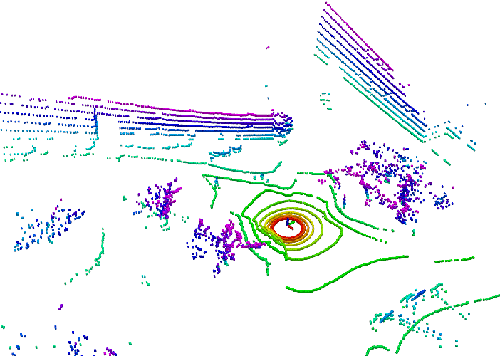}}
	\subfigure[Raw VLP-16 scan 2]{\includegraphics[width=.49\columnwidth]{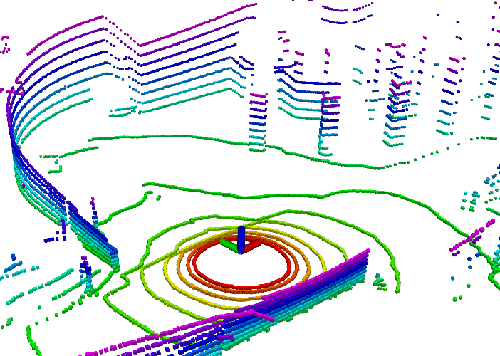}}
	\subfigure[Predicted high-res scan 1]{\includegraphics[width=.49\columnwidth]{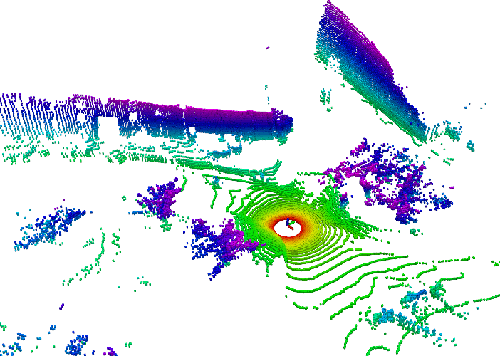}}
	\subfigure[Predicted high-res scan 2]{\includegraphics[width=.49\columnwidth]{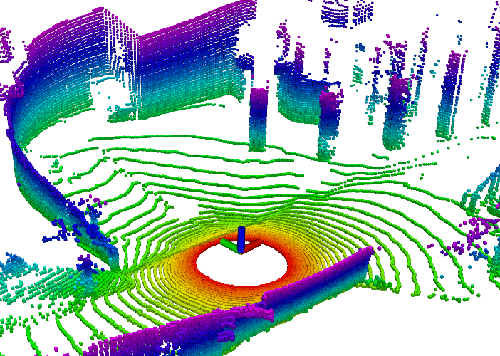}}
	\vspace{-2mm}
	\caption[Lidar super-resolution using Velodyne VLP-16 data.]{Lidar super-resolution using Velodyne VLP-16 data. The low-res point clouds shown in (c) and (d) are captured using a Velodyne VLP-16 lidar. The high-res point clouds shown in (e) and (f) are predicted by our approach (4$\times$ upscaling). Color variation indicates lidar ``ring" index.}
	\label{fig::stevens-demo}
	\vspace{-3mm}
\end{figure}

\section{Stevens dataset}

\begin{figure*}[t]
	\centering
	\subfigure[Satellite image]{\includegraphics[width=.25\textwidth]{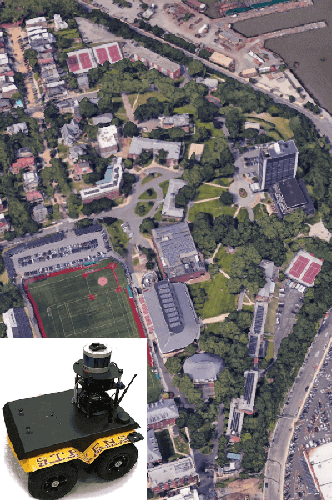}}
	\subfigure[Map using baseline approach]{\includegraphics[width=.25\textwidth]{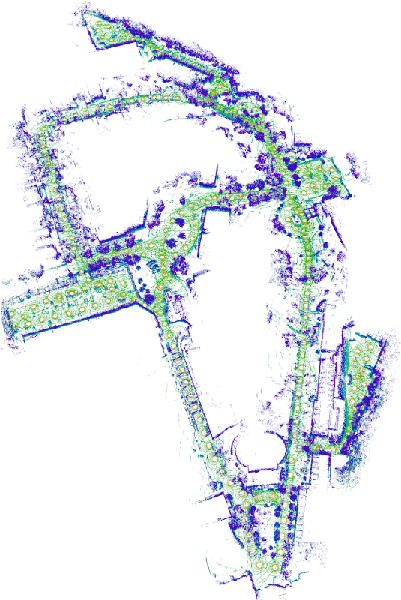}}
	\subfigure[Map using our approach]{\includegraphics[width=.25\textwidth]{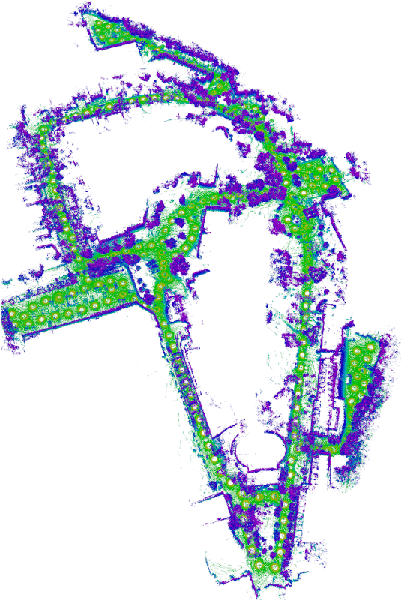}}
	\vspace{-2mm}
	\caption[Lidar super-resolution using Velodyne VLP-16 data.]{Lidar super-resolution using Velodyne VLP-16 data. The low-res point clouds shown in (c) and (d) are captured using a Velodyne VLP-16 lidar. The high-res point clouds shown in (e) and (f) are predicted by our approach (4$\times$ upscaling). Color variation indicates lidar ``ring" index.}
	\label{fig::stevens-map}
\end{figure*}

We show qualitative results from our own dataset, which we refer to as the Stevens dataset, which was collected using a Velodyne VLP-16 mounted on a Clearpath Jackal UGV on the Stevens Institute of Technology campus. This dataset features numerous buildings, trees, roads and sidewalks. Two representative scans from the dataset are shown in Fig. \ref{fig::stevens-demo}.
A satellite photo of the Stevens campus is shown in Fig. \ref{fig::stevens-map}(a). A total number of 278 scans, which are registered with LeGO-LOAM \cite{lego-loam}, are used for generating a 3D map. 
The neural network that is used for testing here is the same as the network that is used in Sec. \ref{sec::gazebo} and \ref{sec::carla}.
Fig. \ref{fig::stevens-map}(b) and (c) show the 3D map created by the raw lidar scans and the inferred lidar scans (4$\times$ upscaling) respectively. The map produced by our inferred scans includes better structural and terrain coverage without using a real high-res lidar.

\clearpage

\end{document}